\documentclass[lettersize,journal]{IEEEtran}

\usepackage{cite}
\usepackage{graphicx}
\usepackage{url}
\usepackage{amsmath}
\usepackage{amssymb}
\usepackage[caption = false, font=footnotesize]{subfig}
\usepackage{float}
\usepackage{booktabs} 
\usepackage{multirow}
\usepackage[colorlinks=true,linkcolor=blue]{hyperref}%
\usepackage{array}
\usepackage{color}
\usepackage{tabularx}
\usepackage{longtable}
\usepackage{tabu}
\usepackage{pifont}
\usepackage{amssymb}
\usepackage{ragged2e}
\usepackage{chemformula}
\begin{document}

\title{Interactive Face Video Coding: A Generative Compression Framework}

\author{ Bolin Chen,~\IEEEmembership{Member,~IEEE}, Zhao Wang, Binzhe Li, Shurun Wang, Shiqi Wang,~\IEEEmembership{Senior Member,~IEEE},  Yan Ye,~\IEEEmembership{Senior Member,~IEEE}
\IEEEcompsocitemizethanks{
\IEEEcompsocthanksitem This work was supported in part by Shenzhen Science and Technology Program under Project JCYJ20220530140816037, in part by Research Grants Council (RGC) General Research Fund 11200323 and 11203220. (Corresponding author: Shiqi Wang)\\
\IEEEcompsocthanksitem Bolin Chen and Binzhe Li are with the Department of Computer Science, City University of Hong Kong, Hong Kong (e-mail: bolinchen3-c@my.cityu.edu.hk, binzheli2-c@my.cityu.edu.hk).\\
\IEEEcompsocthanksitem Zhao Wang is with the School of Computer Science, Peking University, Beijing (e-mail: zhaowang@pku.edu.cn).\\
\IEEEcompsocthanksitem Shurun Wang and Yan Ye are with the DAMO Academy, Alibaba Group (email: shurun.wsr@alibaba-inc.com, yan.ye@alibaba-inc.com).\\
\IEEEcompsocthanksitem Shiqi Wang is with the Department of Computer Science, City University of Hong Kong, Hong Kong, China, and also with the Shenzhen Research Institute, City University of Hong Kong, Shenzhen 518057, China (e-mail: shiqwang@cityu.edu.hk).\\}
}

\maketitle
\begin{abstract}
\justifying
In this paper, we propose a novel framework for Interactive Face Video Coding (IFVC), which allows humans to interact with the intrinsic visual representations instead of the signals. The proposed solution enjoys several distinct advantages, including ultra-compact representation, low delay interaction, and vivid expression/headpose animation. In particular, we propose the Internal Dimension Increase (IDI) based representation, greatly enhancing the fidelity and flexibility in rendering the appearance while maintaining reasonable representation cost. By leveraging strong statistical regularities, the visual signals can be effectively projected into controllable semantics in the three dimensional space (\textit{e.g.,} mouth motion, eye blinking, head rotation, head translation and head location), which are compressed and transmitted. The editable bitstream, which naturally supports the interactivity at the semantic level, can synthesize the face frames via the strong inference ability of the deep generative model. Experimental results have demonstrated the performance superiority and application prospects of our proposed IFVC scheme. In particular, the proposed scheme not only outperforms the state-of-the-art video coding standard Versatile Video Coding (VVC) and the latest generative compression schemes in terms of rate-distortion performance for face videos, but also enables the interactive coding without introducing additional manipulation processes. Furthermore, the proposed framework is expected to shed lights on the future design of the digital human communication in the metaverse. The project page can be found at \url{https://github.com/Berlin0610/Interactive_Face_Video_Coding}.
\end{abstract}


%
\IEEEpeerreviewmaketitle

\vspace{-1.2em} 
\section{Introduction}
\IEEEPARstart{R}{ecent} years have witnessed an exponential increase in the demand for interactive face video coding, along with the development of online video communication in the emerging metaverse. Though substantial progress has been made in video compression~\cite{wiegand2003overview,sullivan2012overview,bross2021overview,lu2019dvc,wang2021Nvidia,6020768}, there are still numerous challenges ahead. First, the high data volume brings barriers to ultra-low delay video communication under low bandwidth conditions. Second, the encoded bitstream, which lacks semantic meanings, does not naturally support direct interactions with raw signals. To tackle these challenges, the interactive face coding framework, which is empowered by ultra-compact, configurable, and semantically meaningful representation, is highly desired. In this manner, the bitstream can be feasibly manipulated at the decoder side based on facial semantics, such that the facial texture, expression and headpose motion can be accordingly reconstructed without introducing additional manipulation processes.

\begin{figure}[tb]
\centering
\subfloat[Face video coding with pre-manipulation at the encoder side]{\includegraphics[width=0.48 \textwidth]{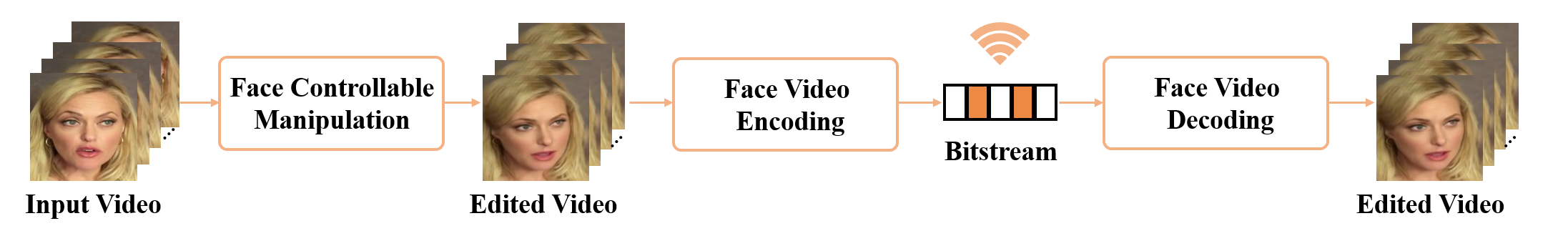}}
\hfil
\subfloat[Face video coding with post-manipulation at the decoder side]{\includegraphics[width=0.48 \textwidth]{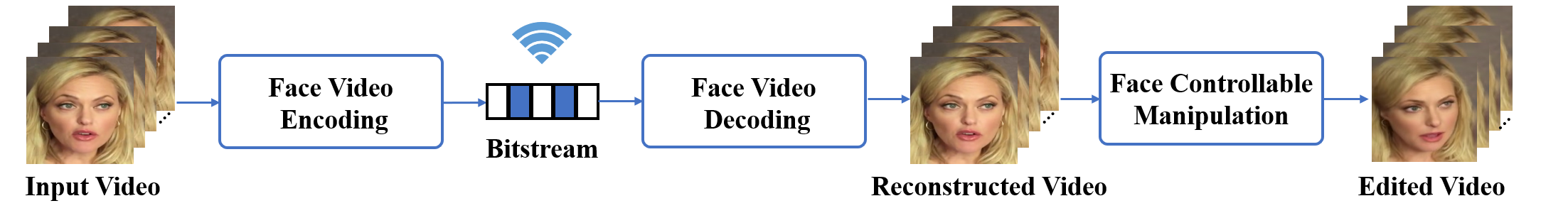}}
\hfil
\subfloat[The proposed IFVC scheme]{\includegraphics[width=0.4\textwidth]{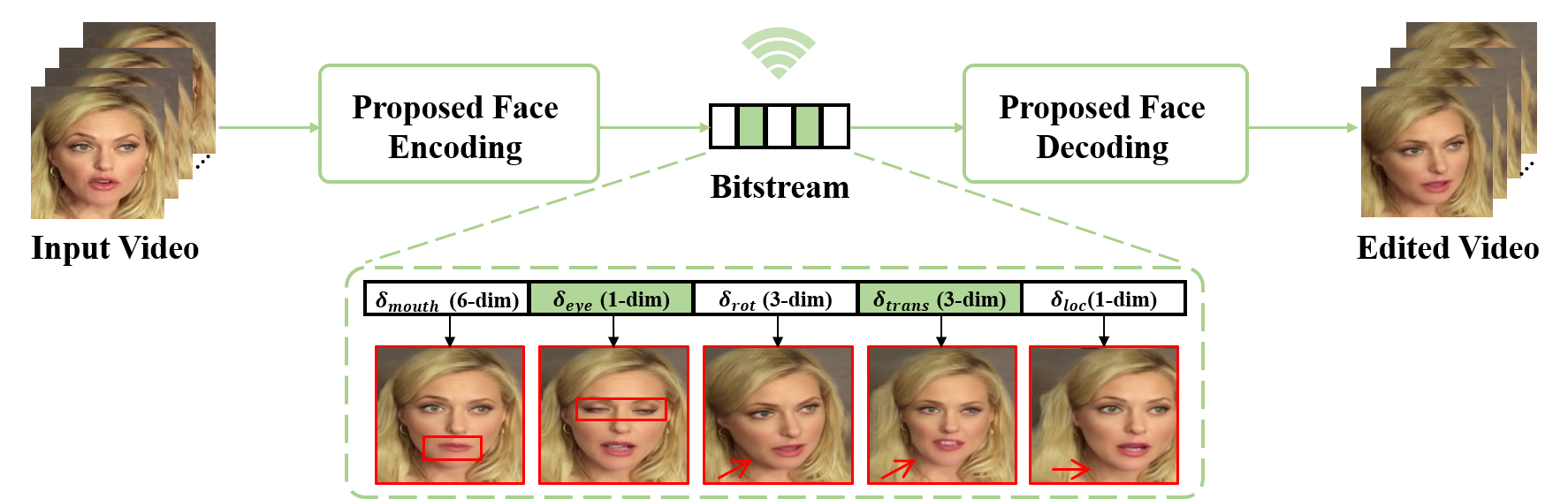}}
\caption{Comparisons of three interactive face video coding paradigms. The interactive face video coding schemes in (a)\&(b) need additional manipulation processes during encoding or decoding face video, whereas (c) shows that our proposed IFVC scheme can directly operate the face coded bitstream featured with ultra compact, configurable and semantically meaningful representations to realize the vivid reconstruction and interaction of face video.}
\vspace{-1.5em} 
\label{framework_comparison}  
\end{figure}

Existing solutions for interactive face video compression  are typically achieved based upon the manipulation on raw pixels, and deep learning has greatly promoted the visual quality of the reconstructed videos. In Fig.~\ref{framework_comparison} (a)\&(b), we provide two typical frameworks that enable the face manipulation in the encoding, transmission and decoding pipeline. In particular, in Fig.~\ref{framework_comparison} (a),
the face video is first controllably manipulated via manipulation algorithms at the encoder side, and the edited face video is further compressed with video compression schemes. Besides, Fig.~\ref{framework_comparison} (b) illustrates that the face video is first compressed by the video compression schemes and the face manipulation is further operated on the reconstructed face video at the decoder side. Regarding the compression of the face videos, numerous methods have been proposed by leveraging the strong regularities on shape, posture and expression variations of the face. In the 1980's, a series of Model-based Coding (MBC) schemes~\cite{1457470,1989Object,150969,364463,lopez1995head,305867,1230217} have been proposed from the perspective of conveying the structural knowledge, aiming at very low bitrate visual communication. 
Hover, it was difficult to reconstruct high-quality talking face images. Recent progress of face reenactment/animation ~\cite{wiles2018x2face,zakharov2019few, FOMM, wang2019few,siarohin2021motion,hong2022depth} based on Generative Adversarial Networks (GAN) ~\cite{goodfellow2014generative} remedies the weaknesses of the traditional MBC technology, fulfilling the promise of talking face compression with a novel paradigm. More specifically, the ultra-low bitrate visual communication frameworks~\cite{facebook2021,ultralow,9859867,CHEN2022DCC,icip2022zhao,10533752,10811831,chen2024beyond,yin2024generative,chen2024standardizing
} can be realized based on an end-to-end animation model, where the motion information (\textit{e.g.,} landmark and keypoint) can be encoded as the transmitted symbols at the encoder side and the high-quality talking face video can be reconstructed at the decoder side. To overcome the fixed viewpoint limitation and realize free-view talking face synthesis, a neural talking-head video synthesis framework (Face\_vid2vid)~\cite{wang2021Nvidia} which decomposes motion-related information and person-specific identity via 3D facial keypoints has been proposed. Though promising rate-distortion (RD) performance has been achieved, Face\_vid2vid lacks the capability in providing freely controllable facial expression (\textit{e.g.,} mouth and eye in interactive coding). As such, it suffers from additional generative processes when interactive manipulation is focused on facial expression such as eye and mouth movements, which may increase the delay and impose additional computational overhead.

To tackle these problems, we propose a new Interactive Face Video Coding (IFVC) framework, allowing the face videos to be effectively encoded to support a variety of interactive demands. The fundamental principle of the framework is enabling the controllable variation of the compact and explainable representation through learnable encoding and decoding, and allowing the interactions at the intrinsic representation level. In contrast to the two existing interactive face coding schemes shown in Fig.~\ref{framework_comparison} (a) $\&$ (b), which are typically challenged by high delay and high complexity, our proposed interactive face coding framework shown in Fig.~\ref{framework_comparison} (c) is capable of directly projecting the face frames into compact embedded facial semantics. In summary, the main contributions of this paper are summarized as follows,

\begin{itemize}
\item{We revisit the IFVC problem, and develop the IFVC framework which attains the good balance among rate, visual quality, degree of freedom in interactivity, and computational complexity. The proposed IFVC framework benefiting from ultra-compact, embedding-configurable, and semantically meaningful representations, is expected to serve as the essential component to warrant the service of immersive video conferencing and metaverse-related activities.}  
\item{We develop the Internal Dimension Increase (IDI) based coding scheme, which converts the input two-dimension (2D) frame to three-dimension (3D) mesh representations internally, and the 3D mesh representation is further manipulated and projected back to the 2D frame at the decoder side. The proposed IDI scheme effectively enhances the flexibility in controllable variation and quality in reconstruction, while maintaining a reasonable representation cost by leveraging the regularities of the face evolution. } 
\item{We design a GAN-based decoder that can facilitate face generation from all dynamics of the intrinsic representation during the interactions. The dense motion flow and facial attention map, which can gain the instantaneous awareness of the dynamics, could make the best use of the deep generative model for producing the desired decoded video. By comparing our proposed scheme with state-of-the-art algorithms in terms of objective and subjective evaluations, it is demonstrated that our method is highly promising in ultra-low bitrate, interactivity enabled and privacy-protected face communications.} 
\end{itemize}

\section{Related Works}
In this section, we review the recent progress of video compression for universal scenarios and specific talking face video based applications (\textit{e.g.,} video conference and live entertainment). Besides, the 3D face generative models are further introduced.
\vspace{-1em}
\subsection{Video Coding for Universal Scenarios}
\subsubsection{Hybrid Video Coding}
Over the past decades,  a series of video coding standards have been developed, including the H.264/Advanced Video Coding (AVC)~\cite{wiegand2003overview}, H.265/High Efficiency Video Coding (HEVC)~\cite{sullivan2012overview} and H.266/Versatile Video Coding (VVC)~\cite{bross2021overview}, greatly enabling a variety of applications including broadcasting, live streaming and video conferencing. Moreover, enabled by deep learning techniques, a variety of video coding tools, such as intra/inter prediction~\cite{zhu2019generative,zhao2018enhanced}, entropy coding~\cite{ma2019convolutional}, interpolation filtering~\cite{liu2019one}, and in-loop filtering~\cite{8630681,8954561}, are further enhanced for better RD performance. However, these block-based hybrid coding schemes designed for universal scenarios still lack the innate-ability in removing redundancies for face videos which typically exhibit strong prior statistical regularities. More importantly, such conventional video codecs cannot support the manipulation over the coding bitstream for immersive interactivity.

\subsubsection{Learning-based Video Coding}
Instead of simply substituting individual modules in the hybrid video coding framework, learning-based image/video compression algorithms optimize the entire compression framework in an end-to-end manner. In particular, the learning-based image compression algorithms~\cite{balle2016end,balle2018variational,DuanLJ0M022,fu2023learned} have shown competitive RD performance compared with JPEG, JPEG2000, HEVC and VVC. In addition, sophisticated learning-based video compression schemes have also been developed. For example, the first end-to-end Deep Video Compression model (DVC)~\cite{lu2019dvc} fully exploited the non-linear ability of neural networks and jointly optimized all the modules for video compression. Besides, Lu \textit{et al.} further presented an online encoder updating scheme~\cite{lu2020content} from the perspective of content adaptation and error propagation. Afterwards, a series of end-to-end video compression algorithms~\cite{hu2021fvc,yang2020learningRLVC,zhao_DPEG,mentzer2021neural,zhang2023elfic,lin2023deepsvc} were put forward to improve the coding performance. More specifically, Hu \textit{et al.}~\cite{hu2021fvc} proposed an end-to-end video compression framework by converting the input video to the latent code representation. The relevant techniques of recurrent learning~\cite{yang2020learningRLVC} and adversarial learning~\cite{zhao_DPEG,mentzer2021neural} were also introduced in the end-to-end compression framework. These learning-based compression algorithms could achieve promising coding efficiency in universal scenes, while there is still room for improvement considering the specific application scenarios such as talking face videos.

\subsection{Model-based Coding for Talking Face}
\subsubsection{Conventional Analysis-synthesis Models}
In the 1980's, the structure information of face images has been exploited in Model-Based Coding (MBC)~\cite{1457470,150969,364463}, realizing low-bandwidth visual communications. Generally speaking, the input face could be economically modelled as symbols including semantic parameters or facial edges based on the analysis model, and subsequently the  decoded symbols can greatly facilitate the reconstruction of face images via the synthesis model. For example, Clark \textit{et al.}~\cite{201932xx} utilized a 3D face model and Chowdhury \textit{et al.}~\cite{305867} employed a generalized cylindrical model as well as a 3D matching technique. In addition, the 3D head model~\cite{lopez1995head} was employed to estimate the 3D head motions for face-specific video compression, and both waveform coding~\cite{836279} and Principal Component Analysis (PCA) models~\cite{koufakis1999very} play important roles in MBC. Moreover, great endeavors have also been made to develop lifelike talking face animation systems~\cite{1230217,4014029}, which aim at achieving automatism, realism and flexibility for interactive services via 3D analysis models and image-based synthesis techniques. However, for these existing MBC techniques, the reconstruction quality of face images was not satisfactory due to handcrafted analysis-synthesis models, thereby limiting the practical applications.  

\subsubsection{2D Representation Based Face Models}
Recently, deep generative models, such as Variational Auto-Encoding (VAE)~\cite{VAE}, GAN~\cite{goodfellow2014generative}, Diffusion Models (DM)~\cite{NEURIPS2021_49ad23d1} have shown great potentials in image and video generation, especially for talking faces. These face synthesis methods~\cite{wiles2018x2face, zakharov2019few, wang2019few, siarohin2021motion,hong2022depth} typically characterize the input frames with latent space~\cite{wang2018every,wang2018recurrent2} (\textit{i.e.,} landmarks or keypoints) and rely on the  powerful inference capabilities of deep generative models to reconstruct these face frames, greatly advancing the quality of experience for ultra-low bitrate visual communication. More specifically, the First Order Motion Model (FOMM)~\cite{FOMM} aims to project face frames into learned 2D keypoints along with their local affine transformations in a self-supervised way. Based on such an end-to-end face animation framework, video chat systems~\cite{facebook2021}~\cite{ultralow}~\cite{9859867} were developed towards ultra-low bit-rate communication. In addition, Feng \textit{et al.}~\cite{9455985} utilized 2D face landmarks and Chen \textit{et al.}~\cite{CHEN2022DCC} adopted compact feature representation as the transmitted animation symbols to reconstruct talking face video. 
Mao \textit{et al.}~\cite{10372532} further exploited StyleGAN priors as hierarchical semantics such that the face coding framework can support machine intelligence and human visual perception in a progressive fashion. Moreover, various novel technologies have been developed or applied to improve the rate-distortion performance, such as spatial-temporal adversarial training~\cite{chen2023csvt}, feature transcoding~\cite{yin2024parametertranslator}, multi-view aggregation~\cite{volokitin2022neural}, residual enhanced coding~\cite{konuko2023predictive}, frame interpolation~\cite{compressing2022bmvc} and multi-reference dynamic prediction~\cite{icip2022zhao}. 
Although these algorithms rely on compact facial parameters in the 2D plane to achieve high-quality face reconstruction, such features cannot ensure highly independent and fully controllable representations, further limiting the flexibility for IFVC.

In view of the limitations of 2D-based talking head representation methods, Wang \textit{et al.}~\cite{wang2021Nvidia} proposed a free-view talking-head video compression framework, addressing the fixed viewpoint limitation and achieving local free-view synthesis by leveraging 3D keypoint representation in a self-supervised learning way. Though impressive performance has been achieved, facial expression information (\textit{i.e.,} eye and mouth motion) cannot be fully separated from these 3D learning-based keypoint perturbations. As a result, the limitations on controllable expression greatly hinder the user interactivity in scenarios such as live entertainment where controlling the facial expressions (\textit{i.e.,} eye and mouth motion) is a main focus. Our proposed IFVC algorithm can achieve controllable face video compression from the perspective of mouth motion, eye blinking, head rotation and head translation based on the 3D face parametric model. By projecting the 2D face frames into compact facial semantics and achieving the representation of 3D facial meshes, the face video can be reconstructed towards high-quality and personalized characterization at ultra-low bitrate.

\begin{figure*}[tb]
\centering
\vspace{-1.5em}
\centerline{\includegraphics[width=0.8 \textwidth]{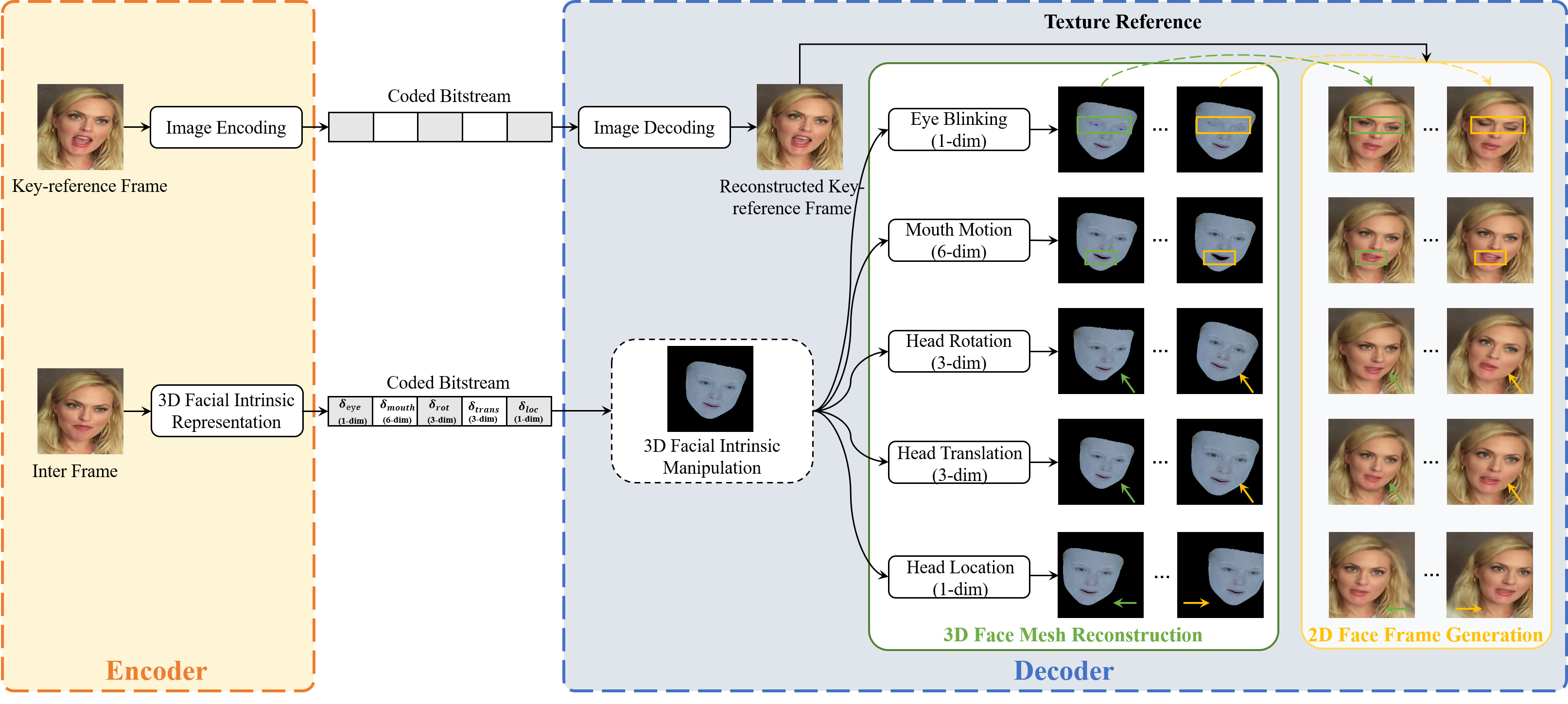}}  
\caption{Illustration of the IDI based coding scheme for realizing the internal conversion between the input 2D face frame and 3D facial mesh representations.}
\label{IDI}
\vspace{-1.2em}
\end{figure*} 

\subsection{3D Face Modeling/Animation with Deep Learning}
\subsubsection{3DMM-based Face Animation}
In 1999, the first 3D Morphable Model (3DMM) model~\cite{Blanz1999AMM} was proposed by Blanz \textit{et al.} for 3D face reconstruction. Afterwards, a series of variations~\cite{ls3dmm,5279762,6654137} have been further developed, based on the traditional principal component analysis (PCA), to characterize low-dimensional semantics from the input face frames, realizing the generation of 3D face meshes. The deep learning based methods have greatly improved the 3D face mesh reconstruction performance by directly regressing 3DMM coefficients from the input and transforming the face template to reconstruct the corresponding face meshes in a supervised~\cite{Zhang2021WeaklySupervisedM3,zhu2020beyond,deng2022fast} or unsupervised~\cite{genova2018unsupervised,wang2021deep,tewari2021learning} manner.

Thanks to such promising 3D face models as well as adversarial learning, the tasks of face reenactment/animation have achieved a great leap towards high-quality generation and controllable manipulation. More specifically, Kim \textit{et al.}~\cite{DVP} first utilized a parametric face model to separate facial parameters (\textit{i.e.,} 3D head position, head rotation, face expression, eye gaze, and eye blinking) from a source actor and further transferred these parameters to another target actor for controllable face reconstruction. Yao \textit{et al.}~\cite{Meshgan} introduced graph convolutional networks to learn the optical flow from the reconstructed 3D meshes for one-shot face synthesis. In addition, Doukas \textit{et al.}~\cite{doukas2020headgan}, Zhang \textit{et al.}~\cite{zhang2021flow}, Ren \textit{et al.}~\cite{9711291} and Yang \textit{et al.}~\cite{Face2FaceRHO} all leveraged the facial 3DMM in the controllable portrait generation from different strategies of multi-modality, multi-task learning or hierarchical motion prediction. Moreover, promising results in 3DMM-based face generative compression have also been shown in~\cite{9810765,9810784}. However, these existing 3DMM generation or compression algorithms are not delicately designed for the IFVC task, commonly faced with challenges such as excessive representation cost for compressing 3D parameters, questionable robustness for face synthesis, and controllable flexibility for facial semantics. In view of these challenges, it is of considerable interest to develop a new IFVC scheme that enjoys the advantages of ultra-low bitrate, high flexibility and vivid reality in face reconstruction and interaction.

\subsubsection{Neural Rendering for Face Animation}

Recently, neural rendering techniques like Neural Radiance Fields (NeRF)~\cite{mildenhall2021nerf} and 3D Gaussian splatting (3DGS)~\cite{kerbl20233d} have shown great potentials in 3D scenes reconstruction with high fidelity and realistic details. Inspired by this, many works~\cite{NEURIPS2022_8cc7e150,li2023hidenerf,chu2024gpavatar,ye2024real3dportrait,Gaussian_Blendshapes,cho2024gaussiantalker,giebenhain2024npga} have explored how to exploit NeRF/3DGS with the deep generative models to achieve realistic and controllable face video animation or reconstruction. In particular, Zeng \textit{et al.}~\cite{NEURIPS2022_8cc7e150} proposed a novel 3D Face Neural Volume Rendering (FNeVR) method for orthogonal adaptive ray-sampling in face signal. Li \textit{et al.}~\cite{li2023hidenerf} proposed a HiDe-NeRF method that could represent the 3D dynamic scene into appearance/motion fields and leverage multi-scale volume features for novel-view synthesis. Besides, to improve the multi-view consistency and reconstruction robustness, Chu \textit{et al.}~\cite{chu2024gpavatar} introduced a dynamic point-based expression field and a multi tri-planes attention mechanism. Different from NeRF-based animation methods, 3DGS-based can greatly improve the rendering speed and achieve real-time reconstruction. Therefore, Ma \textit{et al.}~\cite{Gaussian_Blendshapes} employed 3D Gaussians in the neutral model and expression blendshapes, such that the appearance and motion details of head avatar can be well depicted. Besides, a GaussianTalker model~\cite{cho2024gaussiantalker} was proposed to encode 3DGS attributes into a shared implicit feature representation and deform this with the input driving information for signal reconstruction. Although these neural rendering based animation methods can well achieve faithful identity reconstruction and precise motion control, they are usually faced with poor background generation or solid-color background reference without any texture details.

\begin{figure*}[tb]
\centering
\vspace{-1.5em}
\centerline{\includegraphics[width=0.78 \textwidth]{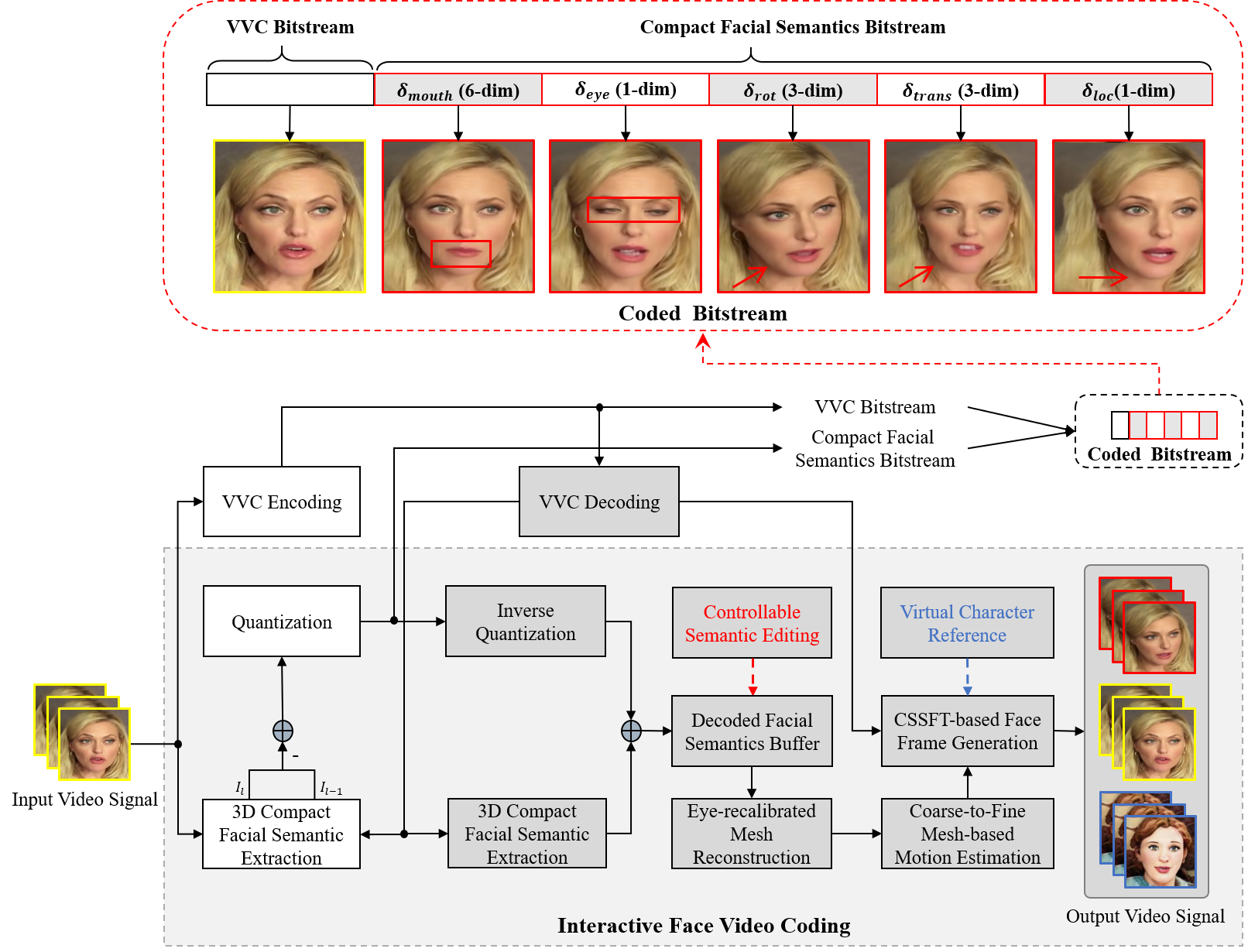}}  
\caption{Illustrations of the encoding and decoding process of IFVC framework, which enables ultra-low bit rate communication (yellow box), metaverse-related interactive functionalities (red box) and user-specified animation/stylization (blue box). The modules of the proposed IFVC decoder are in light gray. }
\label{figframework}
\vspace{-1.2em}
\end{figure*} 

\section{Proposed IDI-based IFVC Framework}
In this section, we give an overall introduction regarding the principle of the proposed IDI-based IFVC framework and its advantages. In addition, the encoding/decoding processes and application scenarios are further discussed in details.
\subsection{Framework Principle and Advantages}
The proposed IFVC framework, which is grounded on the IDI representation, includes the encoder and decoder. As shown in Fig.~\ref{IDI}, at the encoder side, the face video frames are classified as the key-reference frame and inter frames. The key-reference face is encoded by an off-the-shelf image codec (\textit{e.g.,} VVC intra coding), and the inter frames are encoded with the IDI coding scheme. In particular, the IDI coding scheme projects the 2D face into highly-independent parameter space that characterizes the 3D face. The 3D facial intrinsic representations, which are ultra-compact, semantically meaningful (e.g, from the perspectives of mouth motion, eye blinking, head rotation, head translation and head location), and fully controllable, are encoded into the bitstream for inter frame representation. At the decoder side, the manipulation at the bitstream level, instead of the pixel level, enables the intrinsic representation to be reconfigured by request. The 3D mesh can be reconstructed from the varied representation in the 3D parameter space, and 2D face frame is further generated via the reconstructed key-reference frame and the 3D mesh. 

The coded bitstream enjoys the best aspects of compact representation, semantic interpretation, and controllable interaction. There are several advantages for the proposed framework. First, our scheme takes advantage of statistical regularities to a great extent, such that the 14-dimension facial semantic parameter space is sufficient in characterizing the 3D face, facilitating the ultra-low bitrate face video communication. Second, the highly-independent facial semantics based on IDI are interpretable and controllable, and the distinct semantic meanings in terms of mouth motion, eye blinking, head rotation, head translation and head location can benefit a series of interactive applications such as facial semantics editing, immersive interactivity and facial semantics transfer. Third, the reconstruction process based on the editable bitstream ensures user-privacy and high reconstruction quality by synthesizing the 2D face based on the texture template and 3D manipulated face semantics.

\subsection{Overall Encoding/Decoding Processes}
The overall encoding/decoding processes of the proposed IFVC framework is shown in Fig.~\ref{figframework}. Specifically, the input video signal is compressed following the philosophy of predictive coding. In traditional motion-compensated predictive coding, the previously encoded frame is utilized to predict the current frame, and only residuals are encoded. Herein, the prediction is applied on the level of intrinsic visual representations. In particular, the key-reference frame is directly encoded with the state-of-the-art VVC codec, aiming to provide the texture reference for the subsequent generation. The inter frames, which are converted to the 3D facial semantics, are represented in a predictive way by the difference on semantic representations between the current and reconstructed frames. The residuals are finally quantized and entropy coded via a context-adaptive arithmetic coder~\cite{TEUHOLA1978308,1096090}, to ensure the optimal compression performance. 
  
The decoder, the aim of which is reconstructing the face video, is mainly composed of the VVC decoder, 3D semantics decoding, 3D eye-recalibrated mesh reconstruction, coarse-to-fine mesh-based motion estimation and CSSFT-based face frame generation. The key-reference frames is first reconstructed via the VVC decoding and further projected into 3D facial semantics. Subsequently, the facial semantics of inter frames are decoded by context-based entropy decoding and compensation. With the semantics of key-reference and inter frame, the corresponding 3D face meshes can be reconstructed and further sent into the designated mesh-based motion estimation module for estimating the dense motion field and facial attention map. In particular, the reconstruction of 3D face meshes is controllable by modifying the relevant semantic parameters, such that the posture and expression of 3D meshes are varied towards personalized characterization. Finally, given the decoded key-reference frame or a virtual reference character, as well as the explicit facial motion guidance, the talking face video can be reconstructed with high quality thanks to the strong inference capability of the deep generative model. 

\subsection{Application Scenarios}
As shown in Fig.~\ref{figframework}, the proposed IFVC framework enables promising application scenarios, such as ultra-low bit rate communication, metaverse-related interactive functionalities and user-specified animation/stylization, which can be currently unavailable in the latest VVC standard. 

\begin{itemize}
\item{\textbf{Ultra-low Bitrate Communication:} The proposed IFVC framework can characterize the face image with very compact facial semantics, thus enjoying the great advantage of bitrate savings with the same quality video reconstruction compared with the latest VVC codec. As such, our framework can facilitate face video communication such as video conferencing/chat, online broadcasting and live entertainment. In particular, under very tight network bandwidth conditions, our scheme can still ensure that high-quality face video communication is realized where vivid expression, headpose motion and facial texture can be reconstructed.} 
\item{\textbf{Metaverse-related Interactive Functionalities:} The proposed IFVC framework employs the IDI scheme to manipulate the facial embedded semantics and reconstruct the 3D facial mesh toward personalized characterization. In particular, these embedded semantics can be specifically divided into mouth motion, eye blinking, head rotation, head translation and head location, which can be manipulated individually or in combination. As such, it shows a great possibility for future video conferencing and metaverse-related activities that call for friendly interactions and versatile communication. }
\item{\textbf{User-specified Video Animation/Stylization:} The proposed IFVC framework exhibits great robustness when arbitrary face identities are used as texture reference. Given a non-personal face image (\textit{e.g.,} a virtual anime face) as the texture reference, a series of controllable face semantics can animate this reference image to generate face frames with realistic motion and expression. Such a virtual character based video animation/stylization capability can be used to ensure the user privacy in applications such as virtual live-streaming and encrypted video chat. For example, the sender can choose to hide his/her identity, and the controllable face semantics in the proposed system can be used to drive a virtual face and generate faithful motion and expression at the receiver.} 
\end{itemize}

\section{Methodology}
\begin{figure*}[tb]
\centering
\vspace{-1.2em}
\centerline{\includegraphics[width=1 \textwidth]{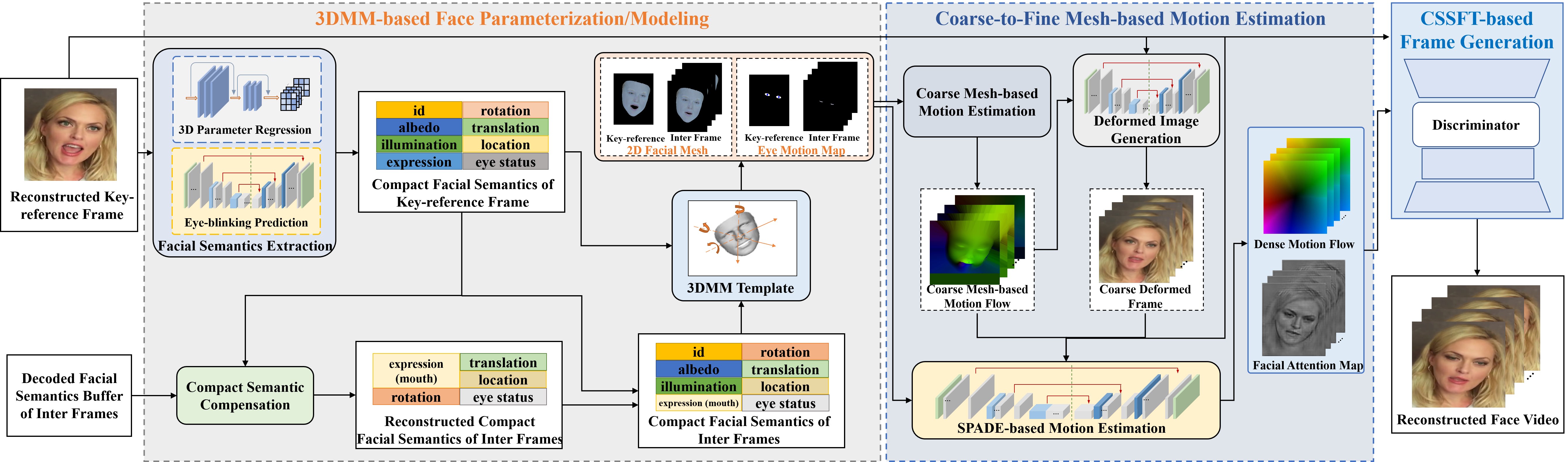}}
\caption{The detailed flowchart of 3DMM-based face parametrization/modeling, coarse-to-fine mesh-based motion estimation and CSSFT-based frame generation.}
\label{fig2}
\vspace{-1.2em}
\end{figure*}

In the traditional image/video coding, the encoder is designed with a high level of flexibility at the expense of encoding complexity, providing sufficient room for optimization. The decoder enjoys the advantage of light-weight decoding, efficiently reconstructing the video sequence given the bitstream. By contrast, the proposed IFVC's encoding/decoding processes, which are fundamentally based on the generative models and manipulatable intrinsic visual representations, can be empowered by the inference capability of the proposed IFVC decoder. As such, at the encoder side, off-the-shelf networks which can meet the application demand are utilized, while the decoder design requires more sophistication. Herein, we will describe each important component/module in the proposed IFVC framework as shown in Fig. \ref{fig2}.

\subsection{3DMM-based Face Parametrization/Modeling} 
Thanks to mature 3DMM techniques, the input high-dimensional face signal could be effectively decomposed into a series of facial semantics and be reconstructed into the corresponding face mesh, but the representation cost of these disentangled semantics is too high and the mesh reconstruction is not robust. In this part, we will introduce our optimization strategy in semantic representation and mesh reconstruction for promising RD performance.

\subsubsection{3D Compact Facial Semantic Extraction}
In IDI based coding, the face frames (i.e., the VVC reconstructed key-reference frame $\hat{K}$, or subsequent inter frames $I_{l} \left (1\le l \le n , l\in Z   \right ) $) are expected to be converted to the semantic representation which is characterized to be compact, semantically meaningful, and controllable. Herein, the pretrained 3D face reconstruction model WM3DR~\cite{Zhang2021WeaklySupervisedM3} is adopted to regress facial semantic representations, including identity coefficients $\delta _{id}\in \mathbb{R}^{80}$, albedo coefficients $\delta _{alb}\in \mathbb{R}^{80}$, scene illumination coefficients $\delta _{illum}\in \mathbb{R}^{27}$, expression coefficients $\delta _{exp}\in \mathbb{R}^{64}$, head rotation coefficients $\delta _{rot}\in \mathbb{R}^{3}$, head translation coefficients $\delta _{trans}\in \mathbb{R}^{3}$ and head location coefficient $\delta _{loc}\in \mathbb{R}^{1}$. The total number of these WM3DR-regressed semantic parameters for each face signal is 258, leading to  high representation cost for compression. Moreover, they also fail to individually decouple mouth movement and eye blinking, resulting in inflexible expression interactions. Finally, the reconstructed mesh based on these semantics also cannot achieve accurate eye movement, thus the fidelity of reconstructed face signal cannot be well guaranteed.

To characterize the high-dimensional face with economical coding bits and robust reconstruction, we further optimize the WM3DR-regressed semantic parameters. First, within the scope of this paper and without loss of generality, we assume the talking face frames from the same sequence share the matched identity $\delta _{id}$, albedo $\delta _{alb}$ and scene illumination $\delta _{illum}$, therefore these coefficients can be directly obtained from the reconstructed key-reference frame without signaling. In addition, according to empirical analysis, the original expression coefficients $\delta _{exp}\in \mathbb{R}^{64}$ are almost used to describe mouth movements. Therefore, we extract the first six dimensions in the 64-dimension expression vector as mouth motion coefficients $\delta_{mouth}\in \mathbb{R}^{6}$. Furthermore, to better depict the blinking intensity of the eye parts, we employ another off-the-shelf facial behavior analysis model OpenFace~\cite{8373812} to predict eye blinking coefficient $\delta _{eye}\in \mathbb{R}^{1}$. As for other facial semantics like head rotation $\delta _{rot}\in \mathbb{R}^{3}$, head translation $\delta _{trans}\in \mathbb{R}^{3}$ and head location $\delta _{loc}\in \mathbb{R}^{1}$, they are so parameter-compact and function-interactive that they can be retained. As such, the final 3D intrinsic visual representations $\delta_{com}$ to be compressed are denoted as, 
\begin{equation}
\label{eq3}
\begin{array}{c}
{
\delta_{com}=\left \{  \delta _{mouth}, \delta _{eye}, \delta _{rot},\delta _{trans}, \delta _{loc} \right \},
}
\end{array}
\end{equation} 
where the total dimension of transmitted compact facial semantics $\delta_{com}$ for each face signal is 14. Such facial embedding strategy brings the benefits of shrinking coding bits, supporting interactivity, and generating vivid face and facial expressions. 

It should be mentioned that the extraction of compact facial semantics both exists at the encoder and decoder sides, where the encoder needs to process all frames and the decoder only extracts the reconstructed key-reference frame. When the extraction of facial semantics (i.e., $\mathcal{\delta}_{com}^{\hat{K}}$ and $ \mathcal{\delta}_{com}^{I_{l}}$) finished at the encoder side, a context-based entropy coding scheme is further employed to realize the high compression efficiency for these semantics. To be specific, they are first inter-predicted between the current and reconstructed frames to remove redundancy. Afterwards, these inter-predicted difference are quantized and further transformed into binary codes based on the principle of zero-order exponential-Golomb algorithm~\cite{TEUHOLA1978308}. Finally, the context-based arithmetic coding model~\cite{1096090} is utilized to generate the final bitstream.

\subsubsection{Eye-recalibrated Mesh Reconstruction}
\label{section: 1}
When the coding bitstream received at the decoder side, the key-reference frame $\hat{K}$ is decoded by the VVC decoder and its corresponding facial semantics $\mathcal{\delta}_{com}^{\hat{K}}$ can be further extracted. Following this, the facial semantics $\hat{\mathcal{\delta}}_{com}^{I_{l}}$ are decoded by the entropy decoding, inverse quantization and semantics compensation. Given these decoded semantic coefficients (\textit{i.e.,} $ \hat{\mathcal{\delta}}_{com}^{I_{l}}$) and other semantic parameters (\textit{i.e.,} $\delta _{id}^{\hat{K}}$, $\delta _{alb}^{\hat{K}}$ and $\delta _{illum}^{\hat{K}}$) extracted from $\hat{K}$, the decoder of WM3DR model~\cite{Zhang2021WeaklySupervisedM3} could synthesize 3D coordinates of face vertices via a parametric 3DMM template. As such, the corresponding 3D face meshes of VVC reconstructed key-reference frame or the subsequent inter frames are given by,
\begin{equation}
\label{eq5}
\begin{array}{c}
{
S =\bar{S} +\delta _{id}^{\hat{K}} \ast S_{id} + \delta _{exp} \ast S_{exp}
},
\end{array}
\end{equation} 
\begin{equation}
\label{eq6}
\begin{array}{c}
{
T =\bar{T} +\delta _{alb}^{\hat{K}} \ast T_{alb} +\delta _{illum}^{\hat{K}} \ast T_{illum}
},
\end{array}
\end{equation} where $S\in \mathbb{R} ^{3\times N} $ and $T \in \mathbb{R} ^{3\times N} $ are the parameterized 3D face shape and texture. In particular, $\bar{S}\in \mathbb{R} ^{3\times N} $ and $\bar{T}\in \mathbb{R} ^{3\times N} $ represent the average neutral shape and texture, where the number of facial vertices $N$ is 35,709. Besides, $S_{id}$, $S_{exp}$, $T_{alb}$ and $T_{illum} $ denote the basis of identity, expression, albebo and scene illumination, respectively. It should be mentioned that $\delta_{exp}$ denotes the expression coefficients $\delta_{exp}^{\hat{K}}$ of the VVC reconstructed key-reference frame, while for the inter frames the first six-dimension expression coefficients of $\delta_{exp}$ are specified by $\hat{\delta}_{mouth}^{I_{l}}$, and the remaining dimensions are compensated by zero padding. 

Based on parameterized 3D face shape $S$ and texture $T$, the 3D face vertices $V$ can be obtained~\cite{Blanz1999AMM}. Subsequently, the corresponding facial mesh $\mathcal{M}$ in the 2D image plane can be transformed as follows,
\begin{equation}
\label{eq7}
\begin{array}{c}
{
\mathcal{M} \Longleftrightarrow  \Psi\left (RV+t  \right ) 
},
\end{array}
\end{equation} where the head rotation matrix $R\in \mathbb{R} ^{3\times 3} $ and the translation vector $t\in \mathbb{R} ^{3} $ can be determined from the corresponding rotation coefficients $\delta _{rot}$ and translation coefficients $\delta _{trans}$. These parameters are from VVC reconstructed key-reference frame $\hat{K}$ or the face semantics of subsequent inter frames $I_{l}$, respectively. Besides, $\Psi$ is an intrinsic matrix of the global camera model that can project 3D information into 2D space.

\begin{figure}[tb]
\centering
\centerline{\includegraphics[width=0.46 \textwidth]{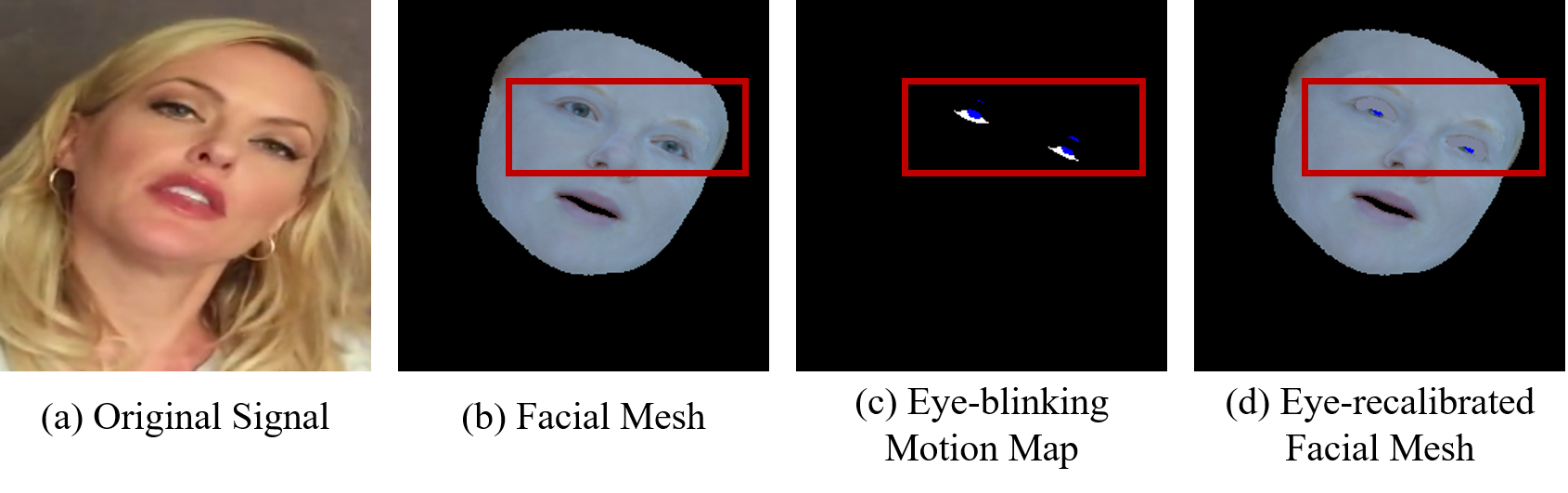}}
\vspace{-1em}
\caption{Visual comparisons of eye-recalibrated facial mesh reconstruction.}
\label{mesh_recalibrate}
\vspace{-1.2em}
\end{figure} 

As illustrated in Fig. \ref{mesh_recalibrate}, the 2D face mesh $\mathcal{M}$ cannot well describe the eye motion status. As such, we introduce the decoded eye-blinking intensity $\hat{\delta} _{eye}$ to re-calibrate the motion of eye regions in $\mathcal{M}$. In principle, the eye regions of talking face frame can be located and extracted via the geometry of transformed 2D face mesh $\mathcal{M}$. Subsequently, the maximum vertical distance between the highest (\textit{i.e.,} $P_{hp}$) and lowest point (\textit{i.e.,} $P_{lp}$) for these marked regions is further adjusted according to the corresponding values of eye-blinking intensities. As a result, a new highest point (\textit{i.e.,} ${P_{hp}}'$) can be defined to re-calibrate these marked regions and obtain the correct eye-blinking areas (\textit{i.e.,} $\mathcal{E}$). This process can be formulated by,
\begin{equation}
\label{eq8}
\begin{array}{c}
{
{P_{hp}}'=P_{lp}-\frac{5-\hat{\delta} _{eye}}{5} \times \left | P_{lp}-P_{hp} \right | 
},
\end{array}
\end{equation} where the range of intensity value $\hat{\delta} _{eye}$ is 0$\sim $5. In particular, for eye-blinking motion, the value of 0 denotes the fully-open status and the value of 5 represents the fully-closed status. 

To conclude, with the decoded semantic parameter set $\hat{\delta}_{com}=\left \{  \hat{\delta} _{mouth}, \hat{\delta}_{eye}, \hat{\delta} _{rot},\hat{\delta}_{trans}, \hat{\delta} _{loc} \right \}$, the desired face motions (\textit{i.e.,} expression and posture) can be accurately represented in the form of 2D facial mesh $\mathcal{M}$ and eye-blinking motion map $\mathcal{E}$.

\begin{figure*}[tb]
\centering
\vspace{-1.2em}
\centerline{\includegraphics[width=1 \textwidth]{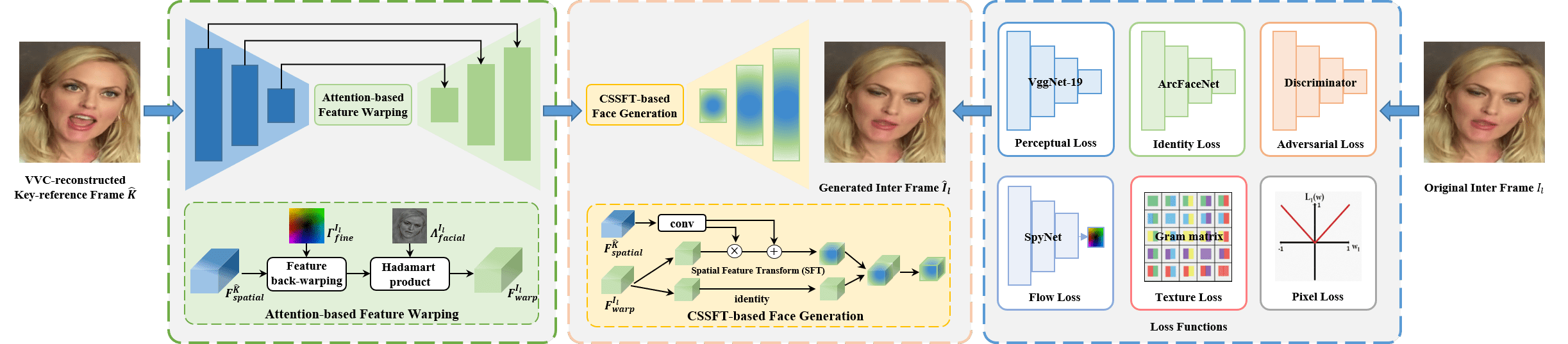}}
\caption{Overview of our proposed CSSFT-based face frame generation module with a series of supervised loss functions.}
\label{fig3}
\vspace{-1.2em}
\end{figure*}

\subsection{Coarse-to-Fine Mesh-based Motion Estimation}
Herein, we first approximate motion of each vertex in the 2D facial mesh from the VVC reconstructed key-reference frame $\mathcal M_{\hat{K}}$ and the current inter frame $\mathcal M_{I_{l}}$. Different from FOMM~\cite{FOMM} and Face\_vid2vid~\cite{wang2021Nvidia} that describe motion from the first order Taylor expansion in the neighbourhood of learned sparse keypoints, our motion representation is dedicated to dense facial meshes instead of 2D frames, and obtained by the difference of each vertex location in $\mathcal M_{\hat{K}}$ with its corresponding location in $\mathcal M_{I_{l}}$. As such, the coarse mesh-based flow $\Gamma^{I_{l}}_{coarse}$ can be estimated as follows,
\begin{equation}
\label{eq9}
\begin{array}{c}
\Gamma^{I_{l}}_{coarse}= \mathbb{G}_{\mathbb{RID}}   \left (  V_{\hat{K}}- V_{{I_{l}}}\right ),
\end{array}
\end{equation} where $V_{\hat{K}}$ and $V_{I_{l}}$ represent each vertex in two facial meshes $\mathcal M_{\hat{K}}$ and $\mathcal M_{I_{l}}$, respectively. $\mathbb{G}_{\mathbb{RID}} \left ( \cdot  \right ) $ means the griddata function used to interpolate the direction change of each mesh point into the 2D plane.

Subsequently, the approximated coarse motion field $\Gamma^{I_{l}}_{coarse}$ and the VVC reconstructed key-reference frame $\hat{K}$ are jointly input into a UNet-like encoder-decoder network~\cite{RFB15a}~\cite{Hourglass2016} for obtaining a coarse deformed frame (\textit{i.e.,} $\mathcal {F}^{I_{l}}_{cdf}$). In particular, $\hat{K}$ is transformed into feature map via the encoder of U-Net architecture and the learned feature map has conducted the feature warping operation with $\Gamma^{I_{l}}_{coarse}$. Finally, the warped feature is further transformed into the coarse deformed frame. The specific process can be described by,
\begin{equation}
\label{eq10}
\begin{array}{c}
{
\mathcal{F}^{I_{l}}_{cdf}=  U_{Dec}\left ( f_{w}\left (U_{Enc}\left ( \hat{K} \right ), \Gamma^{I_{l}}_{coarse} \right ) \right ) 
},
\end{array}
\end{equation} where $U_{Enc}\left ( \cdot \right )$ and $U_{Dec}\left ( \cdot \right )$ represent the feature learning process in the encoder and decoder of UNet architecture, and $f_{w}$ is the back-warping operation. 

The deformed frame $\mathcal {F}^{I_{l}}_{cdf}$ suffers from the absent of eye motion and hair texture, though it could share motion representation similar to the original inter frame  $I_{l}$ from the perspective of posture and expression. This could possibly lead to the unsatisfactory quality in terms of the level of realism in reconstructed face. To address these challenges, we further develop a coarse-to-fine motion estimation scheme via the mesh-approximated coarse motion field $\Gamma^{I_{l}}_{coarse}$, the coarse deformed frame $\mathcal {F}^{I_{l}}_{cdf}$ as well as eye-blinking motion map $\mathcal{E}$. More specifically, we introduce the Spatially-Adaptive Normalization (SPADE) mechanism~\cite{park2019SPADE} (\textit{i.e.,} $\mathbb{S}_{\mathbb{PADE}}\left ( \cdot \right )$) to better preserve semantic information and constantly utilize this semantic information to instruct motion estimation and attention learning. Hence, the dense motion field $\Gamma^{I_{l}}_{fine}$ and facial attention map $\Lambda ^{I_{l}}_{facial}$ can be obtained for realistic face reconstruction as follows,
\begin{equation}
\label{eq11}
\begin{array}{c}
{
\Gamma^{I_{l}}_{fine}=P_{1}\left ( {\small \mathbb{S}_{\mathbb{PADE}}}  \left ( {\small \mathrm{concat} }  \left (  \hat{K}, \mathcal{E}_{\hat{K}}, \mathcal {F}^{I_{l}}_{cdf}, \mathcal{E}_{I_{l}},\Gamma^{I_{l}}_{coarse}\right ) \right )  \right )
},
\end{array}
\end{equation}
\begin{equation}
\label{eq12}
\begin{array}{c}
{
\Lambda ^{I_{l}}_{facial}=P_{2}\left ( {\small \mathbb{S}_{\mathbb{PADE}}}  \left ( {\small \mathrm{concat} }  \left (  \hat{K}, \mathcal{E}_{\hat{K}}, \mathcal {F}^{I_{l}}_{cdf}, \mathcal{E}_{I_{l}},\Gamma^{I_{l}}_{coarse}\right ) \right )  \right )
},
\end{array}
\end{equation} where $P_{1}\left ( \cdot \right )$ and $P_{2}\left ( \cdot \right )$ represent two different predicted outputs, and  ${\small \mathrm{concat} }  \left (\cdot \right ) $ denotes the concatenation operation.

\subsection{CSSFT-based Face Frame Generation}
The strong inference capability of generative adversarial networks greatly benefits the new paradigm of generative face compression. As shown in Fig. \ref{fig3}, the architecture of generative adversarial network is employed to reconstruct high-fidelity talking face video via the motion guidance information. Inspired by the channel-split spatial feature transformation (\textit{i.e.,} CSSFT) mechanism ~\cite{wang2021gfpgan} that can preserve face fidelity for reconstruction, we design a CSSFT-based generative network to animate the VVC reconstructed key-reference frame $\hat{K}$ via dense motion field $\Gamma^{I_{l}}_{fine}$ and facial attention map $\Lambda ^{I_{l}}_{facial}$. First, the VVC reconstructed key-reference frame $\hat{K}$ is fed into the encoder of UNet to obtain multi-scale spatial features $F_{spatial}^{\hat{K}} $. Then, the dense motion field $\Gamma^{I_{l}}_{fine}$ and facial attention map $\Lambda ^{I_{l}}_{facial}$ are further employed to achieve an attention-based feature warping operation for these multi-scale spatial features $F_{spatial}^{\hat{K}} $, thus the warped facial spatial features $F_{warp}^{I_{l}} $ can be obtained,   
\begin{equation}
\label{eq13}
\begin{array}{c}
{
F_{warp}^{I_{l}} =  \Lambda ^{I_{l}}_{facial} \bigodot f_{w}\left ( F_{spatial}^{\hat{K}} , \Gamma^{I_{l}}_{fine} \right ) 
},
\end{array}
\end{equation} where $\bigodot$ represents the Hadamard operation.

The warped result is fed into the CSSFT-based face generation module, where a pair of affine transformation parameters (\textit{i.e.,} scaling $\alpha$ and shifting $\beta$) can be produced from each resolution scale spatial feature $F_{spatial}^{\hat{K}} $ via the convolutional layers. Furthermore, the learned scaling $\alpha$ and shifting $\beta$ parameters are used to modulate the warped facial spatial features $F_{warp}^{I_{l}} $, facilitating to output the transformed facial features. In addition, we concatenate $F_{warp}^{I_{l}} $ and the transformed facial features to perform the face generation ${\hat{I}}_{l}$ that can enjoy the benefits of reconstruction realness and identity. The specific process can be formulated by,
\begin{equation}
\label{eq16}
\begin{array}{c}
{
{\hat{I}}_{l} =  G_{frame} \left ( \mathrm{concat} \left ( F_{warp}^{I_{l}}, \alpha \odot F_{warp}^{I_{l}}+  \beta \right )  \right )
},
\end{array}
\end{equation} where $G_{frame}$ denotes the network layers of the generator that can transform the facial features into face frame. Finally, the multi-scale feature discriminator~\cite{pix2pix2017} is utilized to guarantee that each generated face frame ${\hat{I}}_{l}$ enjoys the realistic reconstruction with the supervision of ground-truth image.

\subsection{Model Optimization}
The self-supervised training strategy is adopted to jointly train the mesh-based motion estimation and CSSFT-based frame generation modules in an end-to-end manner. The objectives in our model training mainly include perceptual loss $\mathcal L_{per}$, adversarial loss $\mathcal L_{adv}$, identity loss $\mathcal L_{id}$, flow loss $\mathcal L_{flow}$, texture loss $\mathcal L_{tex}$ and pixel loss $\mathcal L_{pixel}$. The overall training loss can be described as follows,
\begin{equation}
\label{eq17}
\begin{aligned}
\begin{array}{c}
\mathcal L_{total}=\lambda _{per1}\mathcal L _{per1}+\lambda _{adv1}\mathcal L _{adv1} +\lambda _{per2}\mathcal L _{per2}  \\
+\lambda _{adv2}\mathcal L _{adv2}+ \lambda _{flow}\mathcal L _{flow}+\lambda _{pixel}\mathcal L _{pixel}\\
+\lambda _{tex}\mathcal L_{tex}+\lambda _{id}\mathcal L _{id},
\end{array}
\end{aligned}
\end{equation} where $\mathcal L _{per1}$ and $\mathcal L _{adv1}$ are adopted to ensure the fidelity and naturalness of coarse deformed frame $\mathcal {F}^{I_{l}}_{cdf}$. $\mathcal L_{per2}$, $\mathcal L_{adv2}$, $\mathcal L_{flow}$, $\mathcal L_{pixel}$, $\mathcal L_{tex}$ and $\mathcal L_{id}$ are used to enhance the reconstruction quality of the final generated frames ${\hat{I}}_{l}$. In addition, the hyper-parameters are set as follows: $\lambda _{per1}=10$, $\lambda _{adv1}=1$, $\lambda _{per2}=10$, $\lambda _{adv2}=1$,  $\lambda _{flow}=20$, $\lambda _{pixel}=100$, $\lambda _{tex}=100$  and $\lambda _{id}=40$. 

\section{Experimental Results}
\subsection{Experimental Settings}
\subsubsection{Implementation Details} 
\begin{figure}[t]
\centering
\vspace{-1.2em}
\centerline{\includegraphics[width=0.45\textwidth]{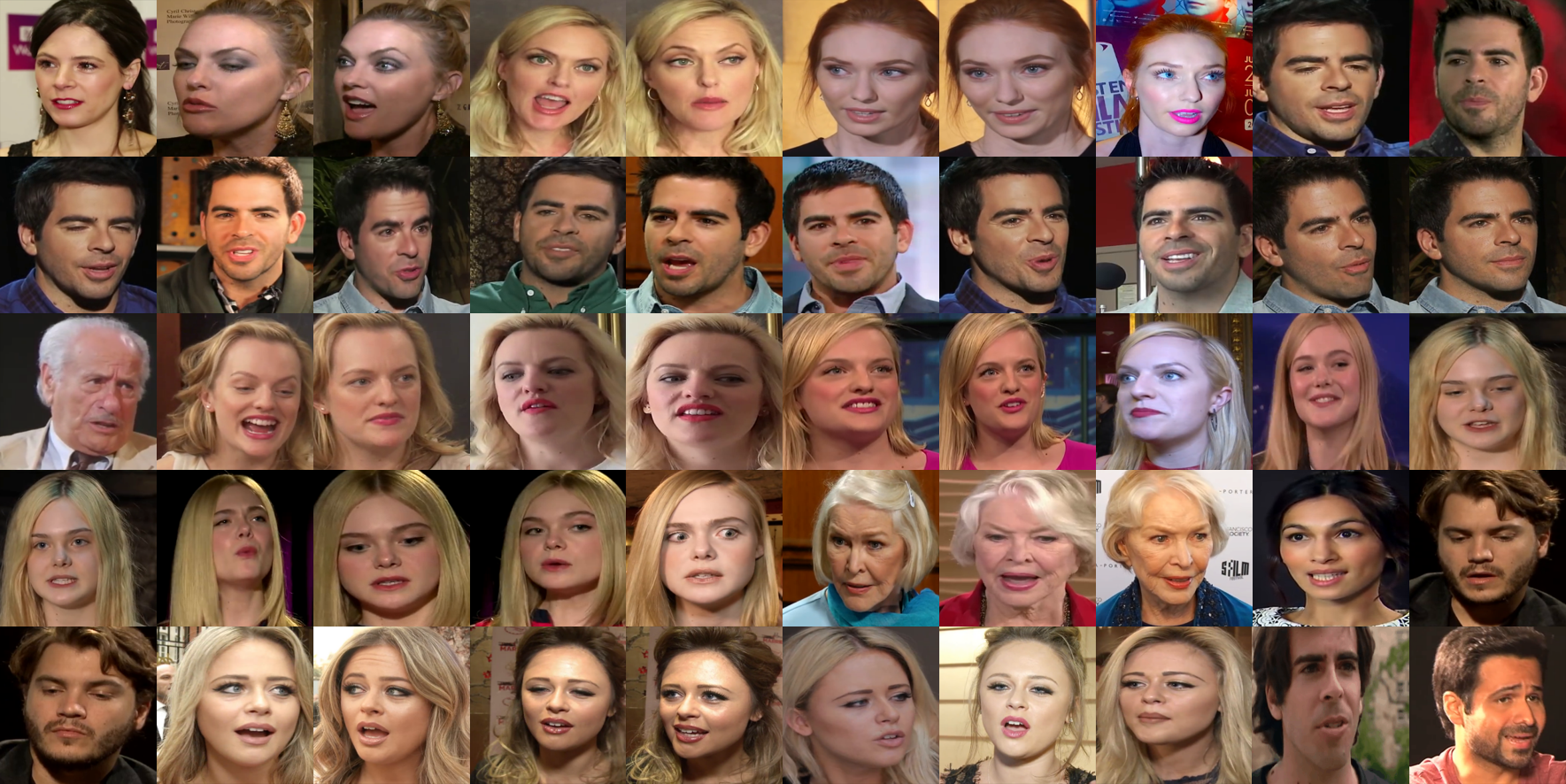}}
\caption{The 50 testing sequences selected and pre-processed from VoxCeleb testing dataset~\cite{Nagrani17}.}
\label{fig4}
\vspace{-1em}
\end{figure}

The proposed generative model is trained based on the widely-used VoxCeleb training dataset~\cite{Nagrani17} with the image size of 256 $\times$ 256 for 130 epochs. Besides, we employ the Adam optimizer with the parameters ($\beta _{1}$ = 0.5 and $\beta _{2}$= 0.999) to train our model, where the network weights are initialized via synchronized BatchNorm. The learning rate is set to 0.0002 and model training is conducted on NVIDIA TESLA-A100 GPUs. For performance evaluation, there are 50 talking face videos selected as our testing sequences from the VoxCeleb testing dataset, as shown in Fig. \ref{fig4}. Each testing sequence contains 250 frames. In addition, the testing virtual characters are sourced from ~\cite{back2021fine}.

\begin{table}[t]
\renewcommand\arraystretch{1.4}
\caption{Average Bit-rate savings of 50 talking face sequences in terms of Rate-DISTS, Rate-LPIPS, Rate-FVD and Rate-MANIQA. ``\#VALUE$!$" means that there are no intersection ranges for two compared curves.}  
\label{table1}
\centering
\resizebox{0.48\textwidth}{!}{
\begin{tabular}{ccccc}
\hline
\multirow{2}{*}{\begin{tabular}[c]{@{}c@{}}Compared Algorithms\\  (Anchor: VVC~\cite{bross2021overview})\end{tabular}} & \multicolumn{4}{c}{Quality Measures}                                          \\ \cline{2-5} 
                                                                                              & Rate-DISTS        & Rate-LPIPS        & Rate-FVD          & Rate-MANIQA       \\ \hline
FOMM (NeurIPS'19)~\cite{FOMM}                                                                             & -35.31\%          & -26.61\%          & -31.03\%          & -35.09\%          \\
FOMM2.0 (CVPR'21)~\cite{siarohin2021motion}                                                                             & -57.30\%          & -55.48\%          & -57.15\%          & -56.87\%          \\
Face\_vid2vid (CVPR'21)~\cite{wang2021Nvidia}                                                                       & -71.81\%          & -68.05\%          & -72.87\%          & -73.95\%          \\
Face2FaceRHO (ECCV'22)~\cite{Face2FaceRHO}                                                                        & -27.59\%          & -7.46\%           & -17.19\%          & -53.11\%           \\
CFTE (DCC'22)~\cite{CHEN2022DCC}                                                                                 & -72.58\%          & -69.70\%          & -70.20\%          & -75.35\%          \\
FNeVR (NeurIPS'22)~\cite{NEURIPS2022_8cc7e150}                                                                            & -38.45\%          & -33.69\%          & -38.45\%          & -45.47\%          \\
HiDeNeRF (CVPR'23)~\cite{li2023hidenerf}                                                                            & 6.46\%            & \#VALUE$!$             & 1.43\%            & -24.79\%          \\
GPAvatar (ICLR'24)~\cite{chu2024gpavatar}                                                                            & -25.13\%          & -16.61\%          & -32.51\%          & \#VALUE$!$           \\ \hline
\textbf{IFVC (Proposed)}                                                                      & \textbf{-75.37\%} & \textbf{-70.29\%} & \textbf{-75.32\%} & \textbf{-80.96\%} \\ \hline
\end{tabular}
}
\vspace{-1.3em}
\end{table}

\begin{figure*}[t]
\centering
\vspace{-2.8em}
\subfloat[Rate-DISTS]{\includegraphics[width=0.3 \textwidth]{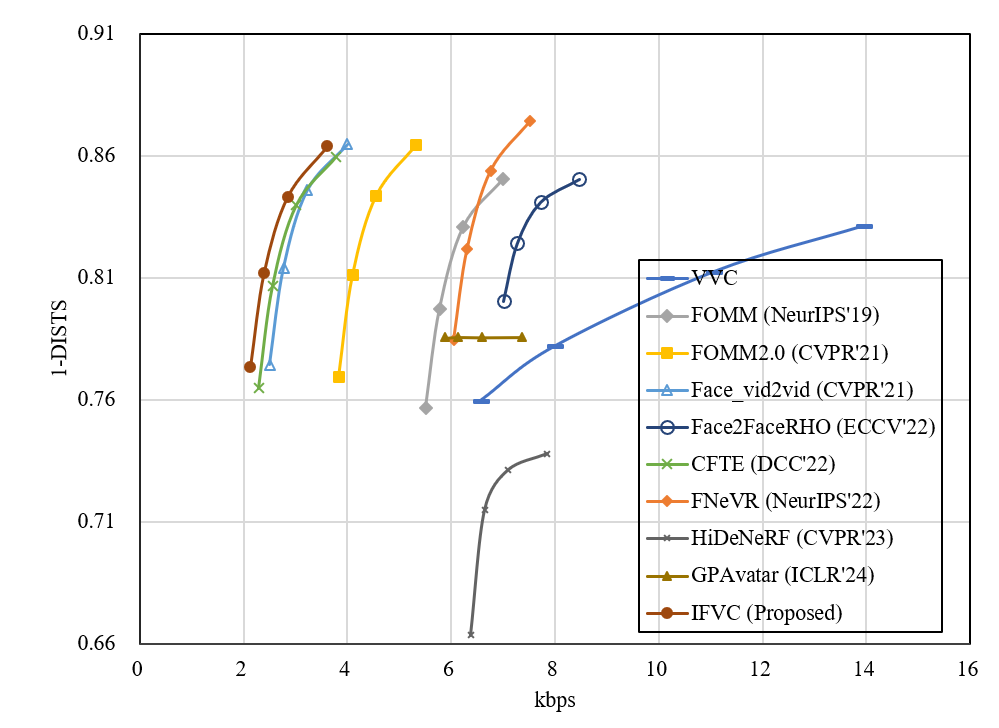}}
\subfloat[Rate-LPIPS]{\includegraphics[width=0.3 \textwidth]{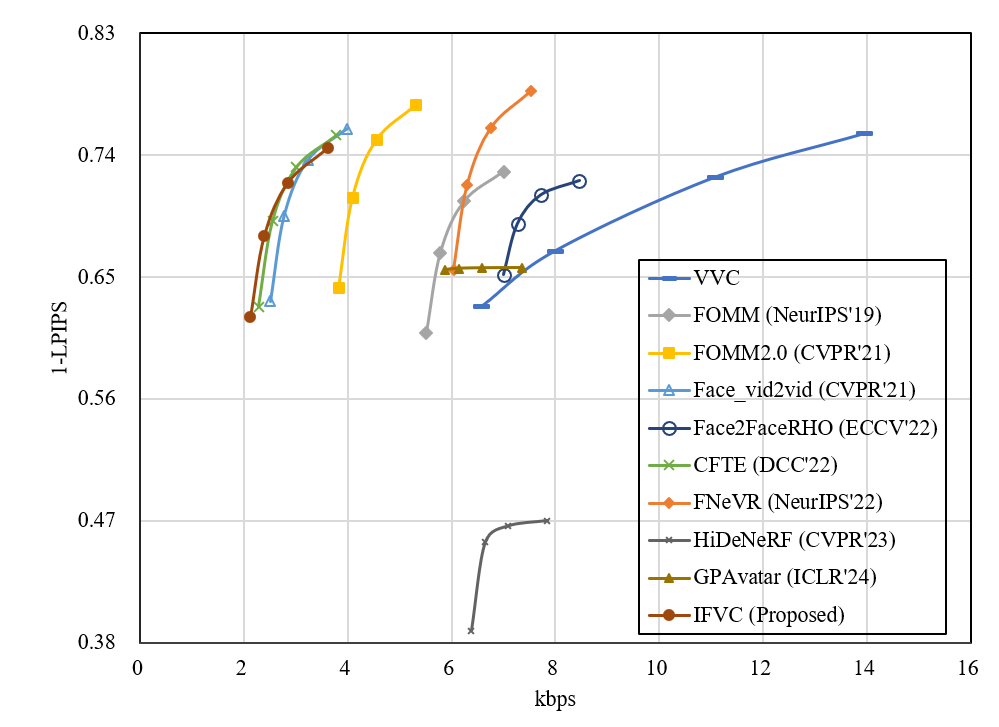}}
\subfloat[Rate-FVD]{\includegraphics[width=0.3 \textwidth]{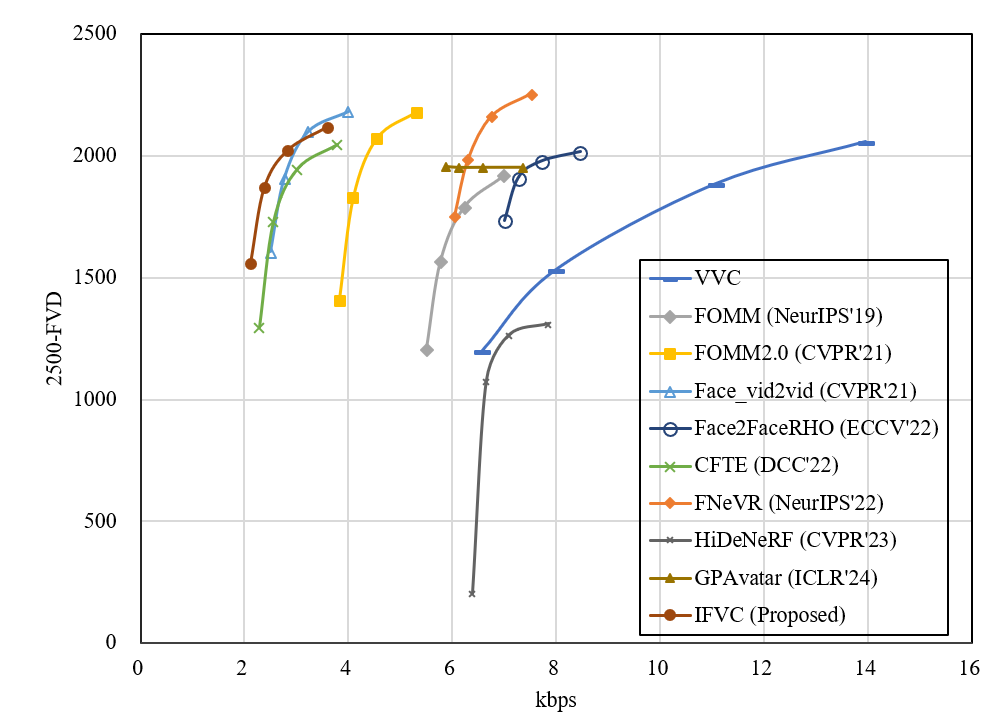}}\\
\vspace{-1em}
\subfloat[Rate-MANIQA]{\includegraphics[width=0.3 \textwidth]{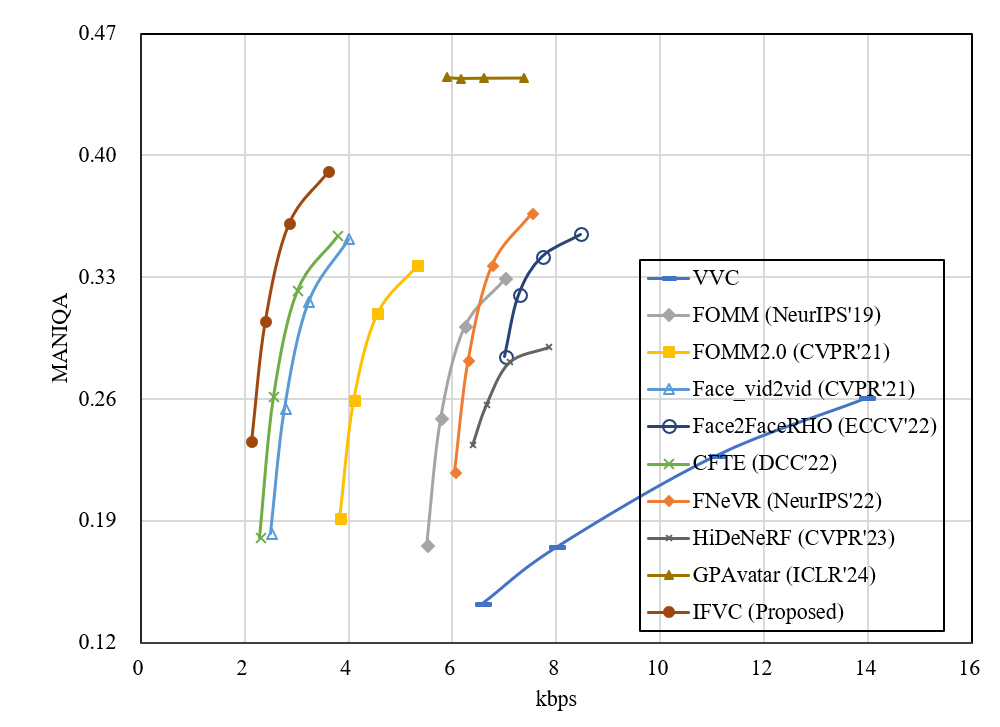}}
\subfloat[Rate-PSNR]{\includegraphics[width=0.3 \textwidth]{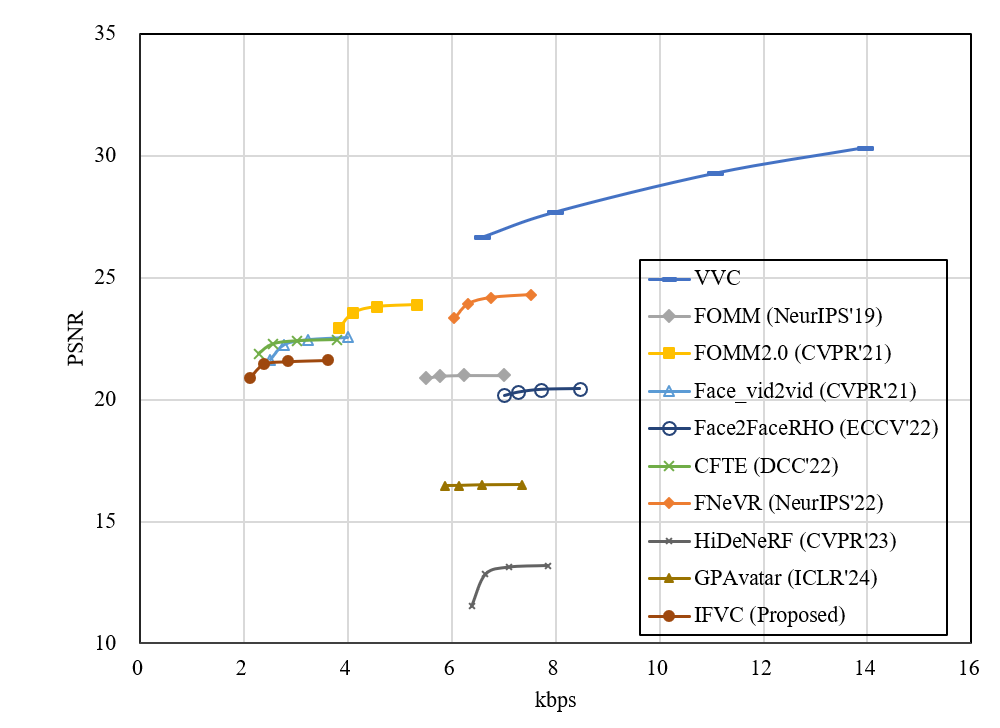}}
\subfloat[Rate-SSIM]{\includegraphics[width=0.3 \textwidth]{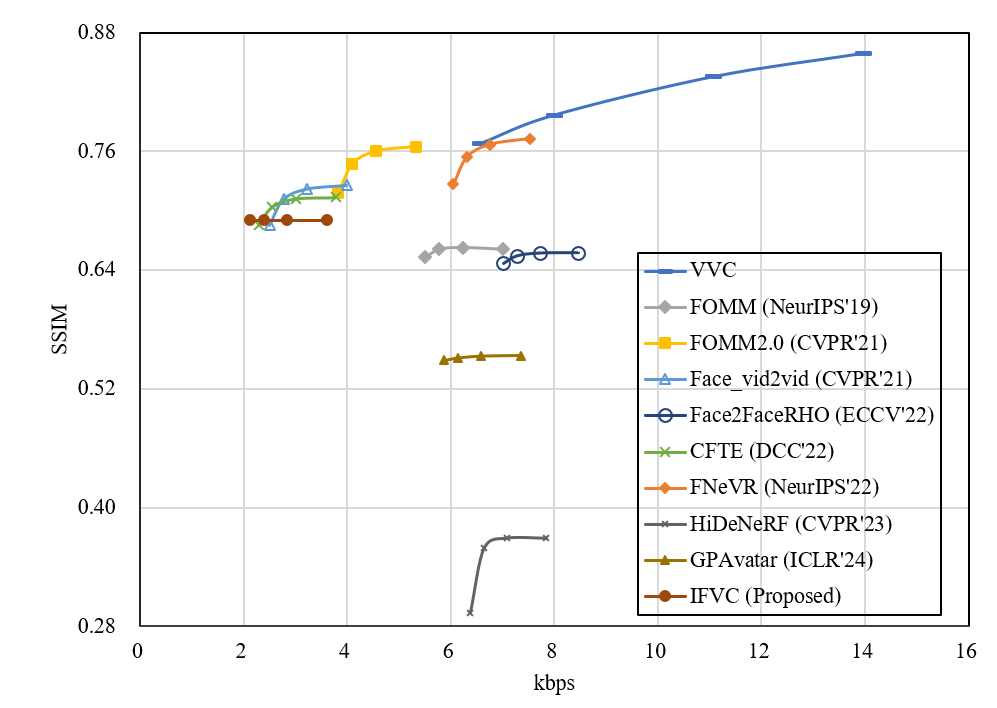}}
\caption{Rate-distortion performance comparisons with VVC~\cite{bross2021overview}, FOMM~\cite{FOMM}, FOMM2.0~\cite{siarohin2021motion}, Face\_vid2vid~\cite{wang2021Nvidia}, Face2FaceRHO~\cite{Face2FaceRHO}, CFTE~\cite{CHEN2022DCC}, FNeVR~\cite{NEURIPS2022_8cc7e150}, HiDeNeRF~\cite{li2023hidenerf} and GPAvatar~\cite{chu2024gpavatar} in terms of DISTS, LPIPS, FVD, MANIQA, PSNR and SSIM.}
\label{fig6}  
\vspace{-0.9em}
\end{figure*}

\begin{figure}[tb]
\centering
\vspace{-0.3em}
\subfloat[Mouth Motion]{\includegraphics[width=0.4 \textwidth]{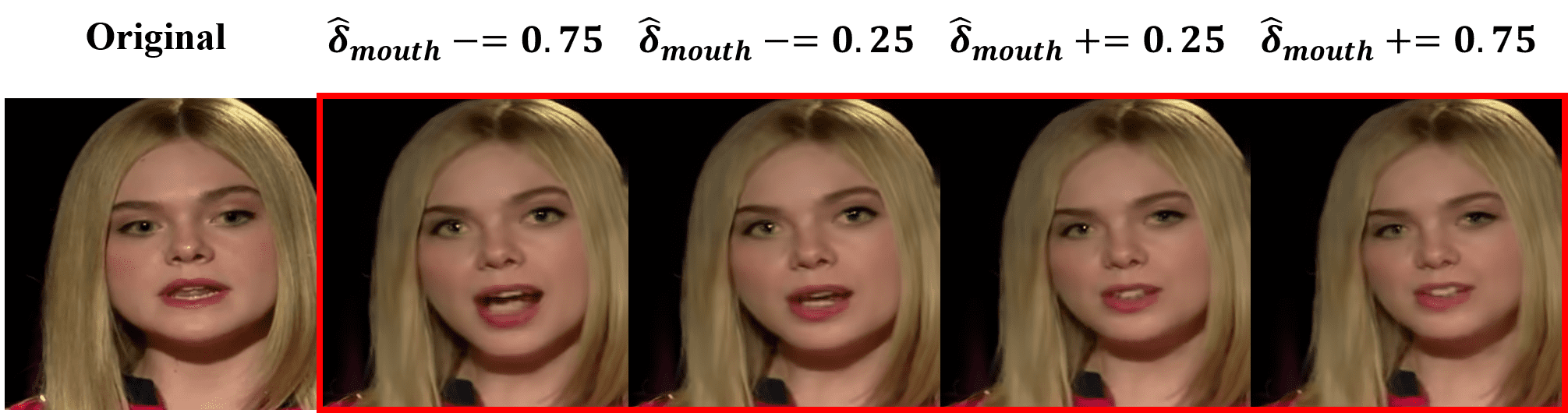}}
\vspace{-0.9em}
\subfloat[Eye Blinking]{\includegraphics[width=0.4 \textwidth]{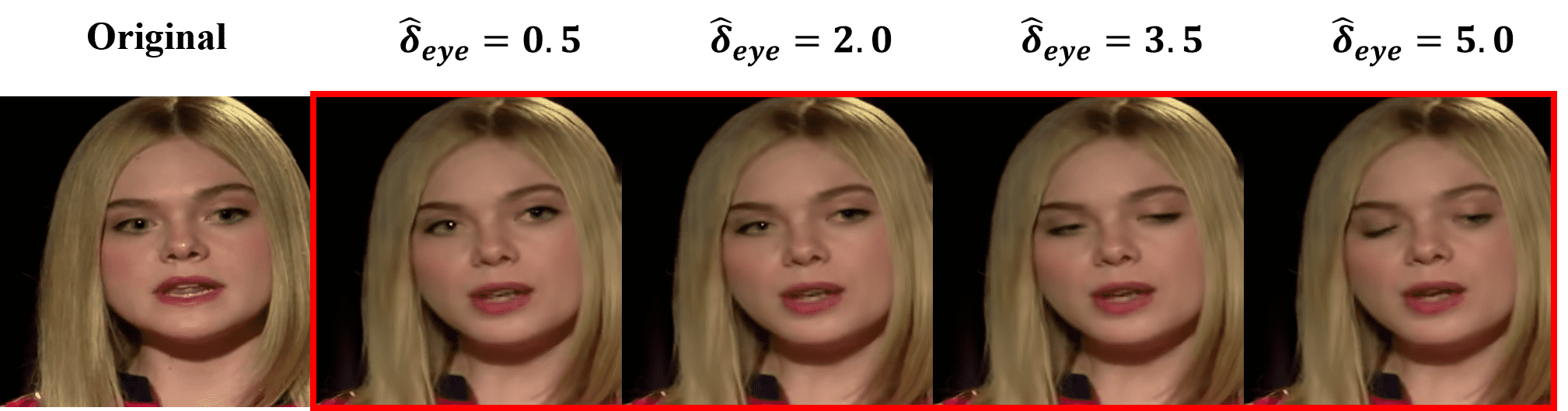}}
\vspace{-0.9em}
\subfloat[Head Rotation]{\includegraphics[width=0.4 \textwidth]{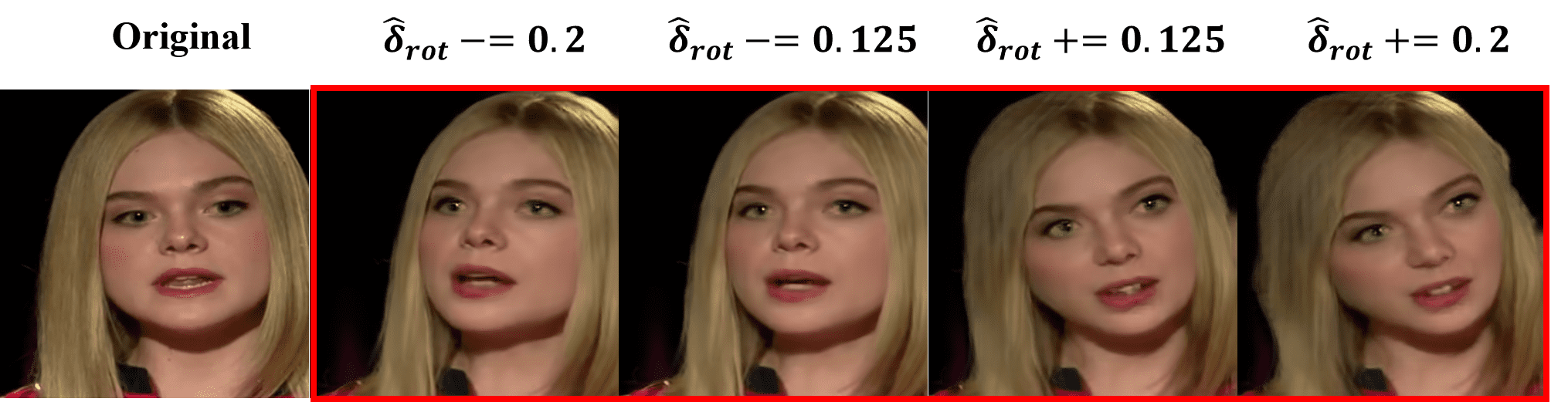}}
\vspace{-0.9em}
\subfloat[Head Translation]{\includegraphics[width=0.4 \textwidth]{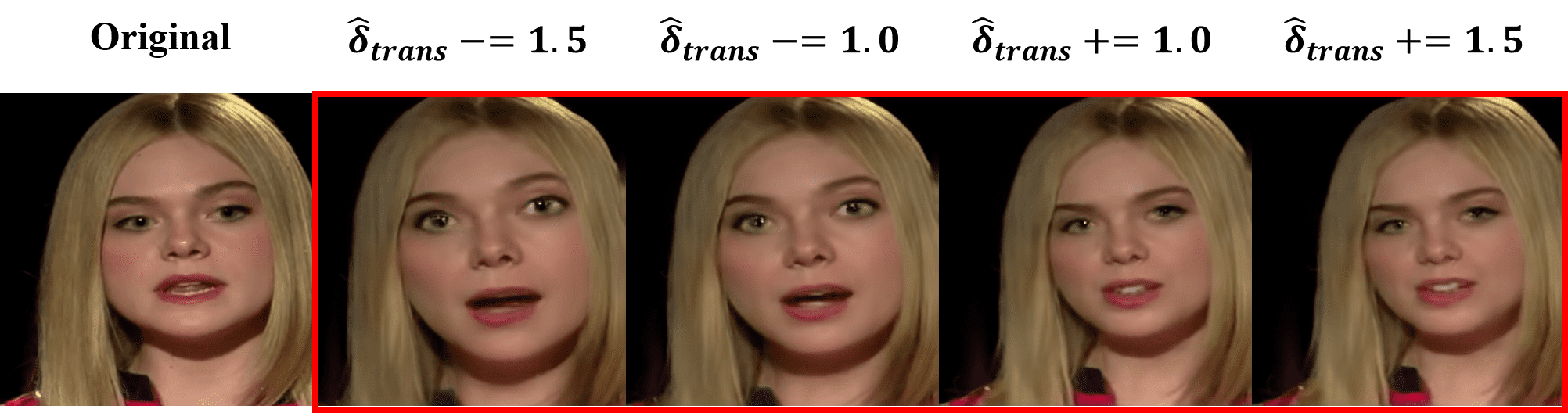}}
\vspace{-0.9em}
\subfloat[Head Location]{\includegraphics[width=0.4 \textwidth]{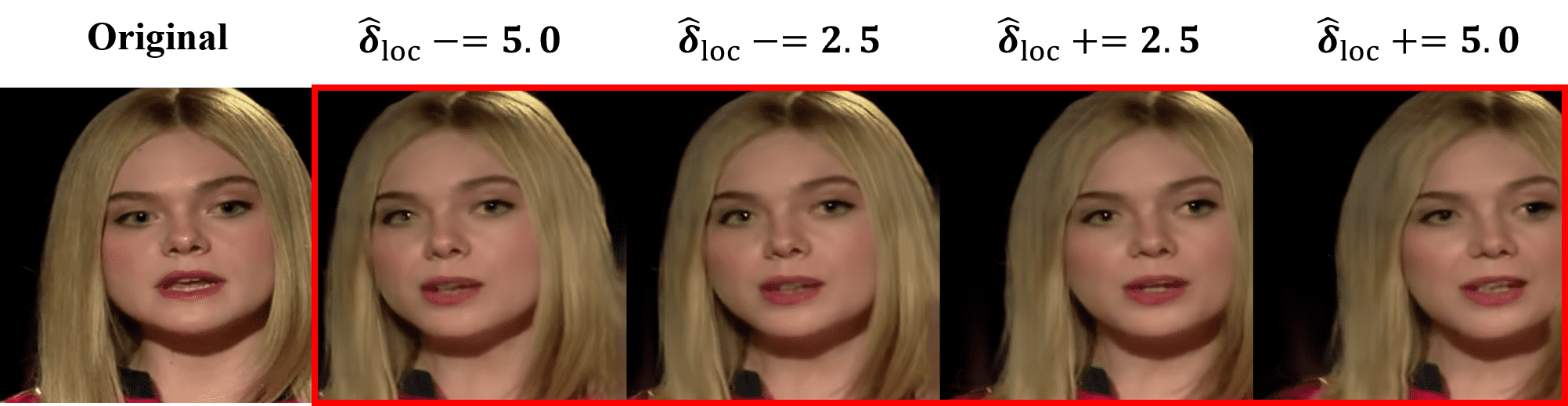}}
\caption{Examples on interactive face coding in terms of individual facial semantic and various interactive degrees. More video examples can be found in \href{https://github.com/Berlin0610/Interactive_Face_Video_Coding}{project page}.} 
\label{fig8}  
\vspace{-2em}
\end{figure}

\begin{figure}[tb]
\centering
\vspace{-1.8em}
\subfloat[Sequence 10]{\includegraphics[width=0.24 \textwidth]{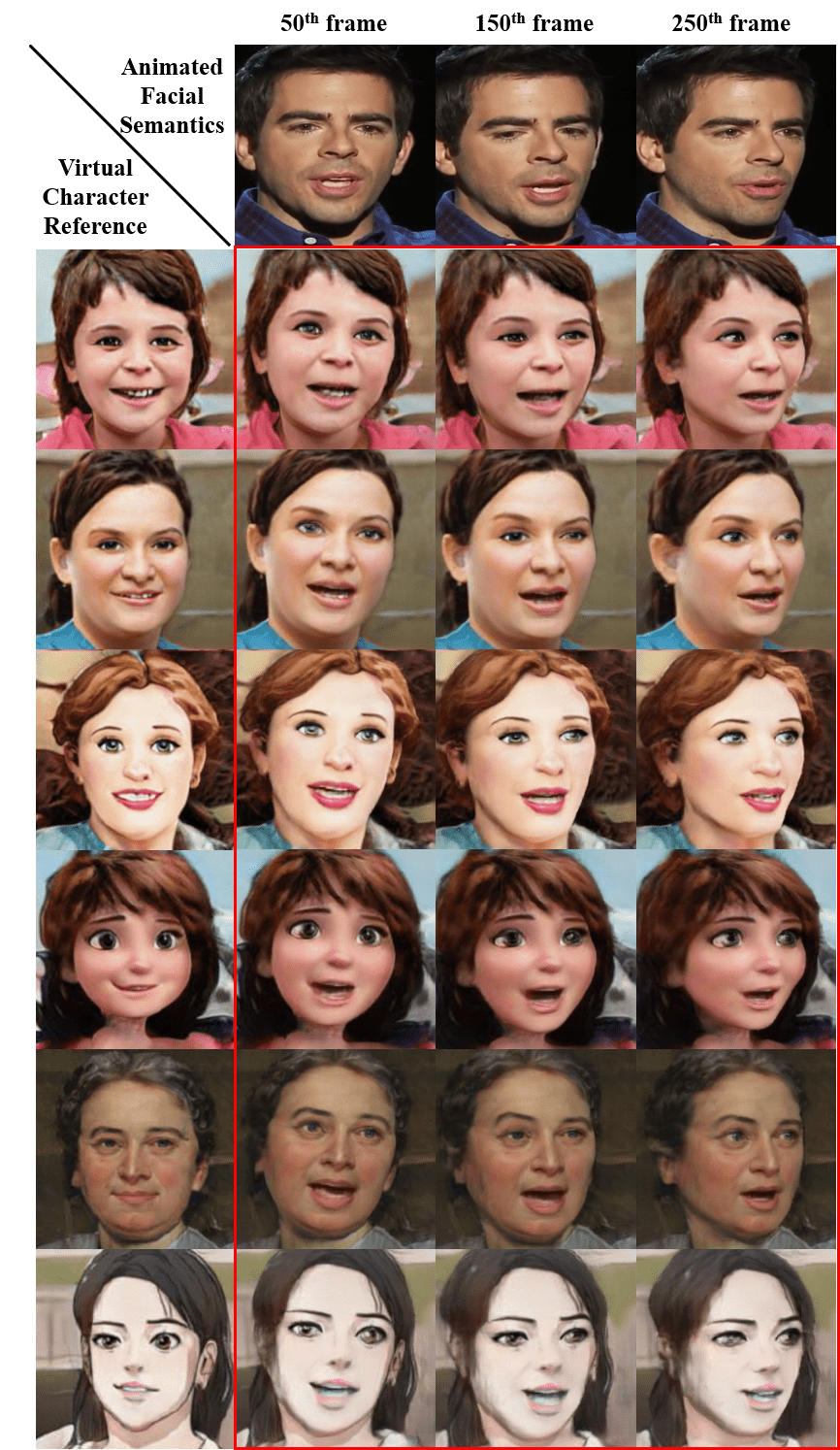}}
\hfil
\subfloat[Sequence 30]{\includegraphics[width=0.24 \textwidth]{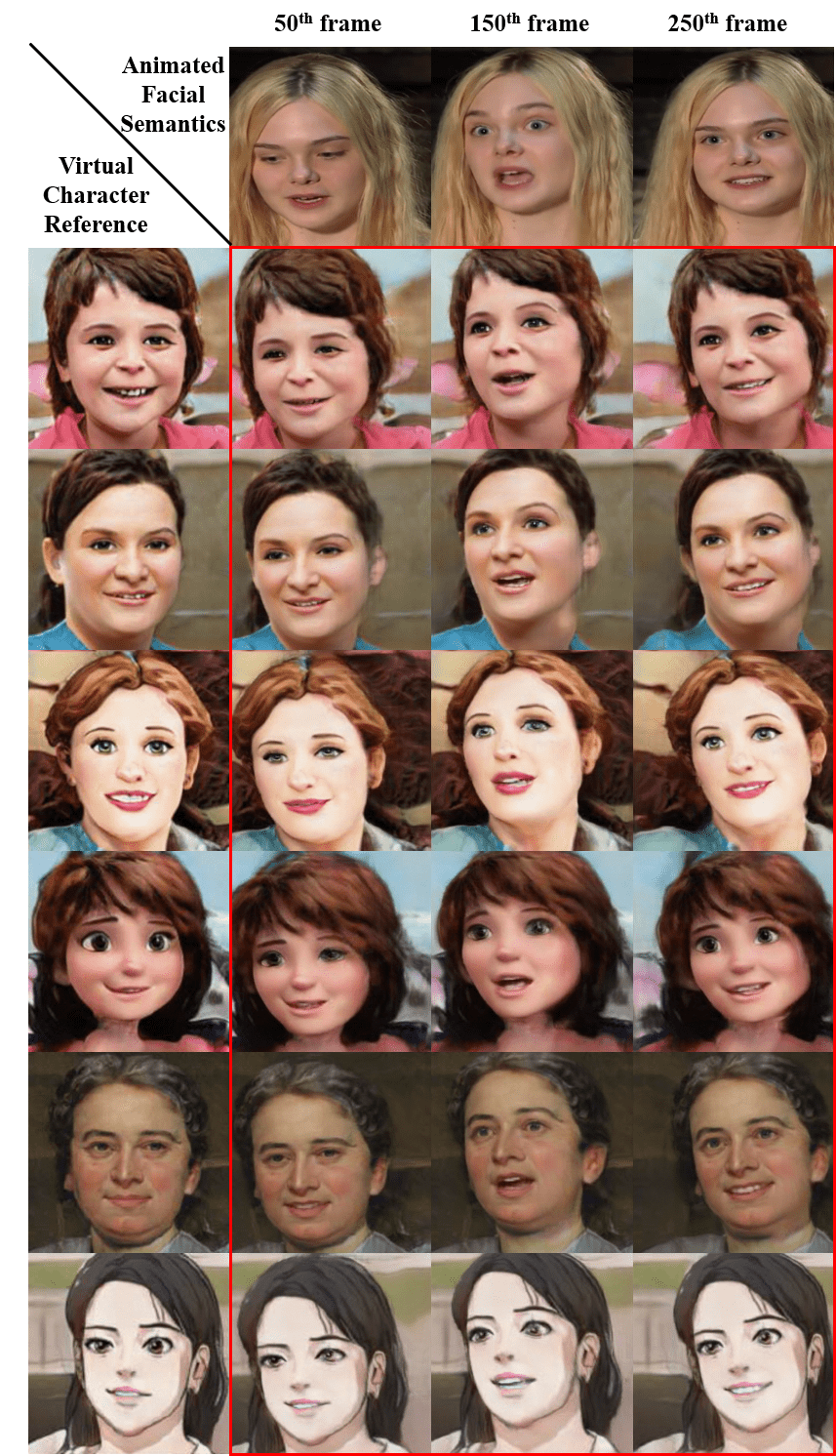}}
\hfil
\caption{Examples on animating virtual character reference with facial semantics. More video examples can be found in \href{https://github.com/Berlin0610/Interactive_Face_Video_Coding}{project page}.
}
\label{fig9}  
\vspace{-0.5em}
\end{figure}

\begin{figure*}[!t]
\centering
\vspace{-2.5em}
\subfloat[Sequence 2]{\includegraphics[width=0.32 \textwidth]{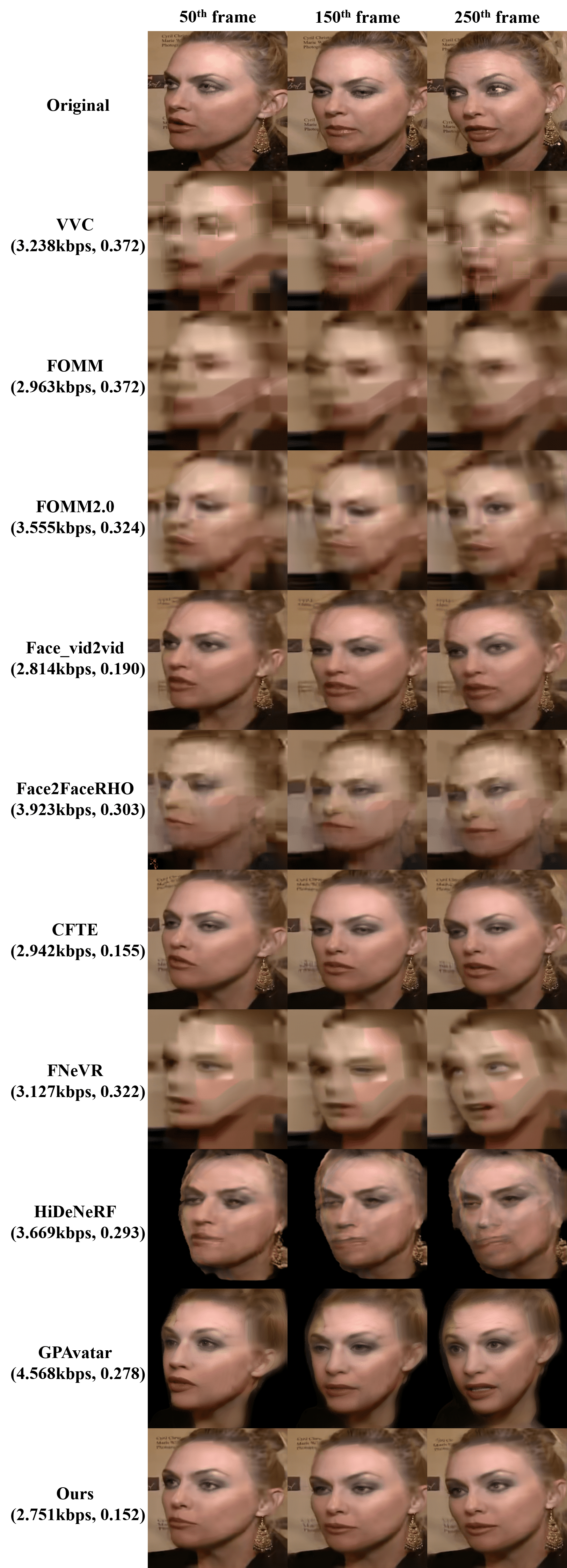}}
\hfil
\subfloat[Sequence 20]{\includegraphics[width=0.32 \textwidth]{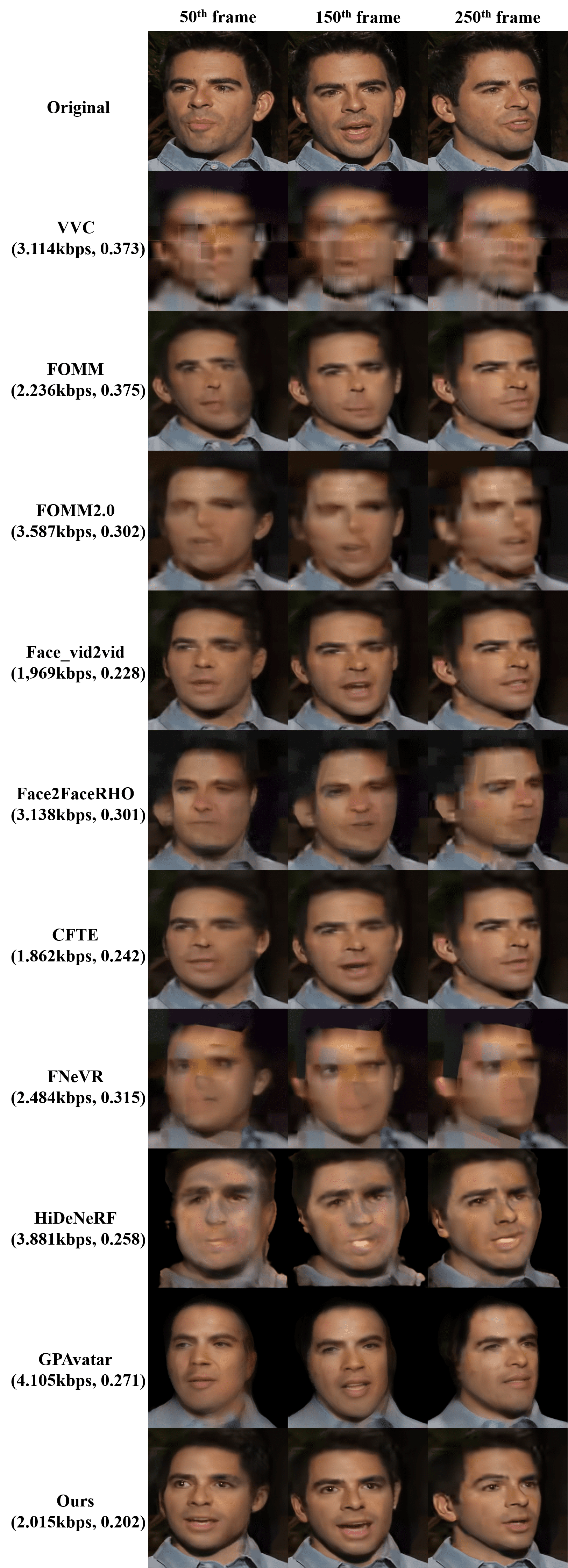}}
\hfil
\subfloat[Sequence 33]{\includegraphics[width=0.32 \textwidth]{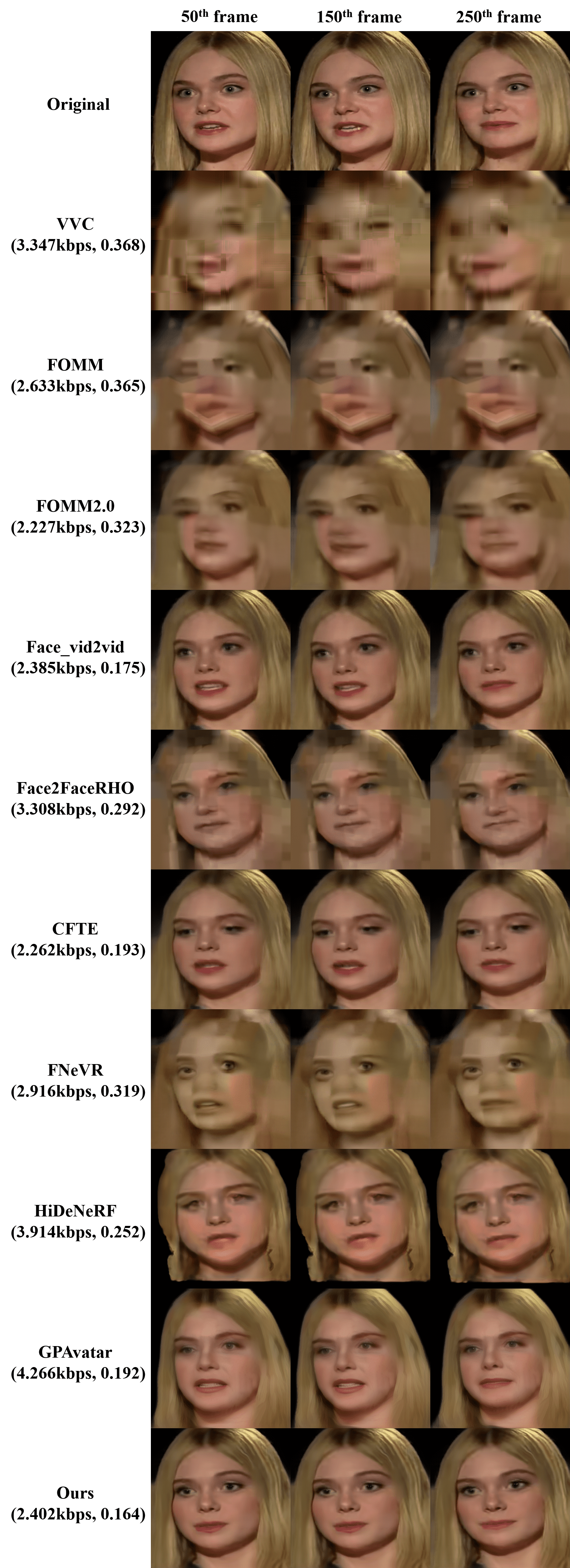}}
\caption{Visual quality comparisons among VVC~\cite{bross2021overview}, FOMM~\cite{FOMM}, FOMM2.0~\cite{siarohin2021motion}, Face\_vid2vid~\cite{wang2021Nvidia}, Face2FaceRHO~\cite{Face2FaceRHO}, CFTE~\cite{CHEN2022DCC}, FNeVR~\cite{NEURIPS2022_8cc7e150}, HiDeNeRF~\cite{li2023hidenerf}, GPAvatar~\cite{chu2024gpavatar} and Ours at the similar bit rate. The values on the left represent coding bits and DISTS value, where lower DISTS value indicates better quality. More video examples can be found in \href{https://github.com/Berlin0610/Interactive_Face_Video_Coding}{project page}.
}
\label{fig7}
\vspace{-1em}
\end{figure*}

\subsubsection{Comparison Methods}
To validate the performance, we compare our proposed IFVC scheme with the latest hybrid video coding standard VVC (VTM10.0)~\cite{bross2021overview} and eight generative compression schemes, including FOMM (NeurIPS'19)~\cite{FOMM}, FOMM2.0 (CVPR'21) ~\cite{siarohin2021motion}, Face\_vid2vid (CVPR'21)~\cite{wang2021Nvidia}, Face2FaceRHO (ECCV'22)~\cite{Face2FaceRHO}, CFTE (DCC'22)~\cite{CHEN2022DCC}, FNeVR (NeurIPS'22)~\cite{NEURIPS2022_8cc7e150}, HiDeNeRF (CVPR'23)~\cite{li2023hidenerf} and GPAvatar (ICLR'24)~\cite{chu2024gpavatar}.
In particular, we adopt the Low-Delay-Bidirectional (LDB) configuration in VTM 10.0 reference software for VVC, where the quantization parameters (QP) are set to 45, 47, 50 and 52. As for other generative compression algorithms, they are rooted in GAN-based image animation models and could be further  migrated into the encoder-decoder-based coding framework. It should be mentioned that in addition to compact feature representations with different parameter dimensions, the test configurations of these algorithms are exactly the same as the proposed IFVC framework. In particular, the key-reference frame is compressed via the VTM10.0 with the QPs of 37, 42, 47 and 52, and compact parameters of other subsequent frames are compressed via a context-adaptive arithmetic coder.

\subsubsection{Evaluation Measures}
To comprehensively evaluate the quality of reconstructed face video, we adopt four learning-based quality measures (i.e., Deep Image Structure and Texture Similarity (DISTS)~\cite{dists}, Learned Perceptual Image Patch Similarity (LPIPS)~\cite{lpips}, Fréchet Video Distance (FVD)~\cite{Unterthiner2019FVDAN} and Multi-dimension Attention Network for no-reference Image Quality Assessment (MANIQA)~\cite{yang2022maniqa}) and two traditional quality measures (i.e., Peak Signal-to-Noise Ratio (PSNR)~\cite{2009Mean} and Structural Similarity Index Measure (SSIM)~\cite{wang2004image}). These measures mainly involve pixel-level, perceptual-level, temporal consistency and no-reference evaluations. Smaller scores of DISTS/LPIPS/FVD indicate higher perceived image quality, and vice versa. As for MANIQA/PSNR/SSIM, higher scores represent better perceived quality. In addition to quality scores, the coding bits of transmitted bitstream are also summarized. As such, rate-distortion curve (\textit{i.e.,} RD-curve) and Bjøntegaard-delta-rate (\textit{i.e.,} BD-rate), which have been the commonly-used evaluation methods in video compression task, can be adopted to evaluate the compression performance.

\vspace{-0.8em}
\subsection{Performance Comparisons}
\subsubsection{Rate-Distortion Performance}
Fig. \ref{fig6} demonstrates the rate-distortion performance of our proposed compression scheme compared with VVC and eight generative compression schemes in terms of different-domain quality measures. More specifically, compared with the state-of-the-art video coding standard VVC, significant bitrate savings have been achieved regarding the learning-based quality measures. In addition, when using the FOMM, FOMM2.0 and Face2FaceRHO algorithms as anchors, the advantage of our IFVC scheme is still evident. Compared with the latest Face\_vid2vid and CFTE compression algorithms, there are still performance gains in terms of DISTS/FVD/MANIQA, and marginal loss has been perceived when using LPIPS as the quality measure. In comparisons with NeRF-based compression algorithms like FNeVR/HiDeNeRF/GPAvatar, our proposed IFVC algorithm can achieve obvious RD performance improvement in terms of DISTS/LPIPS/FVD/MANIQA.

It should be mentioned that both the proposed IFVC and other generative compression methods cannot show any quality evaluation advantages in pixel-level alignment. It can be attributed that generative compression algorithms mainly reconstruct face signal in perceptual-level feature domain via the strong inference capabilities of deep generative models. As such, although these algorithms can achieve realistic visual reconstruction, traditional objective quality measures like PSNR/SSIM designed for the pixel-level distortion calculation may not be appropriate for their quality evaluation~\cite{compressionmeasure,10477607}.

Table \ref{table1} illustrates that our proposed IFVC scheme is able to achieve advantageous bit-rate savings for 50 testing sequences in terms of DISTS, LPIPS, FVD and MANIQA. In particular, our proposed scheme can achieve 75.37\% bit-rate savings in terms of DISTS, 70.29\% bit-rate savings in terms of LPIPS, 75.32\% bit-rate savings in terms of FVD and 80.96\% bit-rate savings in terms of FVD in comparison with the latest VVC codec. Besides, the proposed compression scheme is also superior to the state-of-the-art generative compression algorithms.

\begin{figure*}[t]
\centering
\vspace{-1.2em}
\subfloat[Architecture Modules]{\includegraphics[width=1 \textwidth]{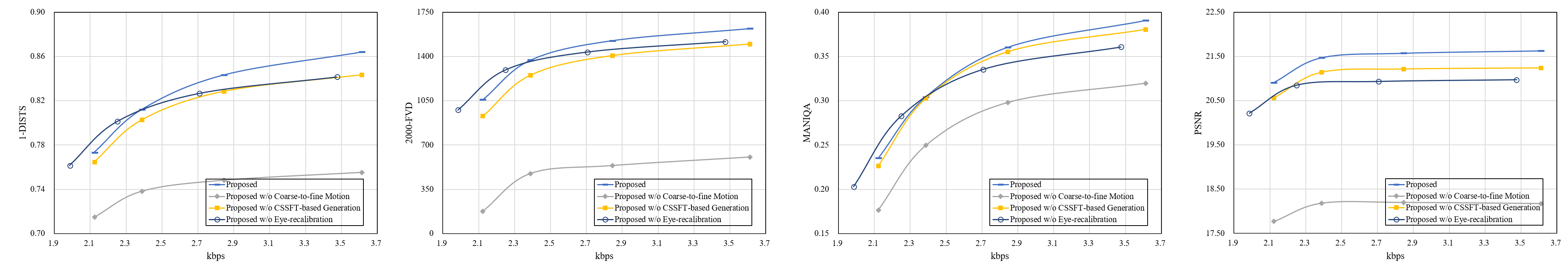}}\\
\vspace{-0.8em}
\subfloat[Mouth Parameter Selection]{\includegraphics[width=1 \textwidth]{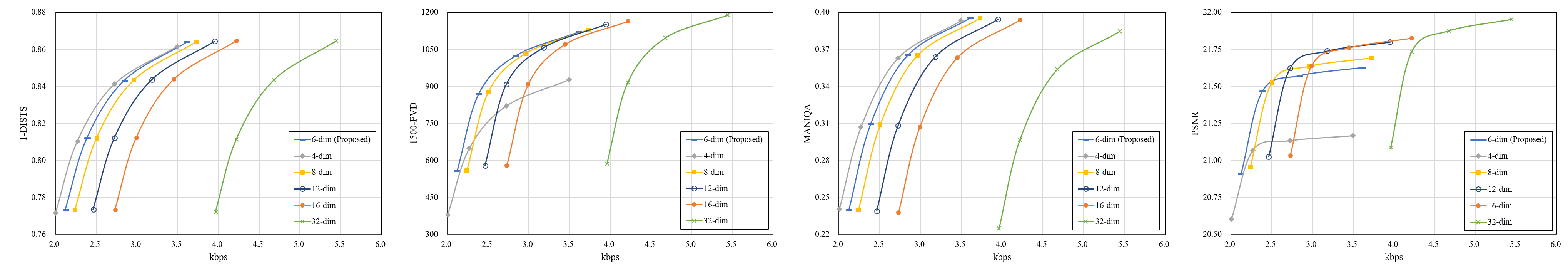}}\\
\vspace{-0.8em}
\subfloat[Optimization Functions]{\includegraphics[width=1 \textwidth]{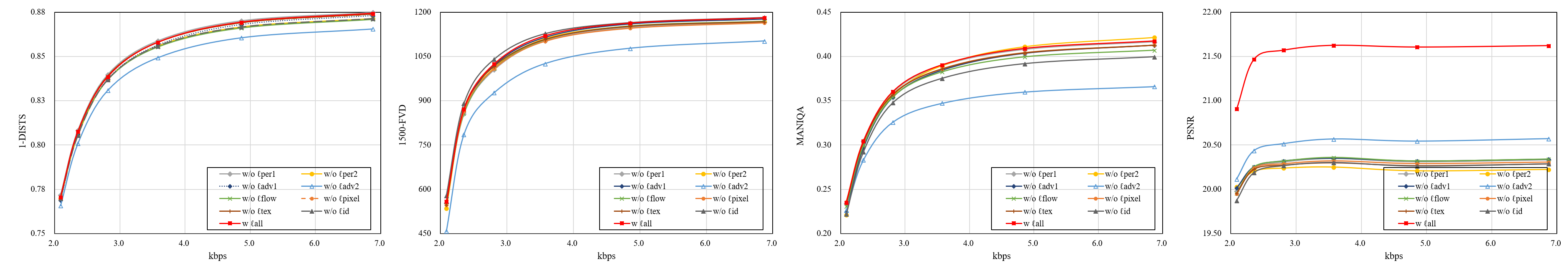}}
\caption{Ablation study in architecture modules, mouth parameter selection and optimization functions in terms of DISTS, FVD, MANIQA and PSNR. }
\label{ablation}  
\vspace{-0.9em}
\end{figure*}

\begin{table*}[tb]
\centering
\renewcommand\arraystretch{1.1}
\caption{User preference in 2AFC comparison at similar coding bits consumption. The average bitrate, DISTS, LPIPS, FVD, MANIQA, PSNR and SSIM of 15 sequences are provided.} 
\label{table2}
\begin{tabular}{cllllllll}
\hline
Algorithms           & \multicolumn{1}{c}{\begin{tabular}[c]{@{}c@{}}Bitrate\\  (kbps) \end{tabular}} & \multicolumn{1}{c}{\begin{tabular}[c]{@{}c@{}}DISTS\\  ($\downarrow$)\end{tabular}} & \multicolumn{1}{c}{\begin{tabular}[c]{@{}c@{}}LPIPS \\ ($\downarrow$)\end{tabular}} & \multicolumn{1}{c}{\begin{tabular}[c]{@{}c@{}}FVD \\ ($\downarrow$)\end{tabular}} & \multicolumn{1}{c}{\begin{tabular}[c]{@{}c@{}}MANIQA\\  ($\uparrow$)\end{tabular}} & \multicolumn{1}{c}{\begin{tabular}[c]{@{}c@{}}PNSR\\  ($\uparrow$)\end{tabular}} & \multicolumn{1}{c}{\begin{tabular}[c]{@{}c@{}}SSIM\\  ($\uparrow$)\end{tabular}} & \multicolumn{1}{c}{\textbf{\begin{tabular}[c]{@{}c@{}}Preference\\  (\%)\end{tabular}}} \\ \hline
VVC~\cite{bross2021overview} / Ours           & 3.50 / 2.72                                                                   & 0.36 / 0.16                                                              & 0.58 / 0.27                                                              & 3365.88 / 473.96                                                       & 0.08 / 0.37                                                               & 22.44 / 21.91                                                           & 0.67 / 0.71                                                             & \textbf{0.0 / 100.0}                                                                    \\
FOMM~\cite{FOMM} / Ours          & 2.83 / 2.72                                                                   & 0.37 / 0.16                                                              & 0.57 / 0.27                                                              & 2818.10 / 473.96                                                       & 0.09 / 0.37                                                               & 19.99 / 21.91                                                           & 0.62 / 0.71                                                             & \textbf{0.0 / 100.0}                                                                    \\
FOMM2.0~\cite{siarohin2021motion} / Ours       & 3.53 / 2.72                                                                   & 0.32 / 0.16                                                              & 0.49 / 0.27                                                              & 2167.16 / 473.96                                                       & 0.11 / 0.37                                                               & 21.83 / 21.91                                                           & 0.67 / 0.71                                                             & \textbf{0.0 / 100.0}                                                                    \\
Face\_vid2vid~\cite{wang2021Nvidia} / Ours & 2.64 / 2.72                                                                   & 0.18 / 0.16                                                              & 0.30 / 0.27                                                              & 549.98 / 473.96                                                        & 0.26 / 0.37                                                               & 22.63 / 21.91                                                           & 0.73 / 0.71                                                             & \textbf{0.0 / 100.0}                                                                    \\
Face2FaceRHO~\cite{Face2FaceRHO} / Ours  & 3.80 / 2.72                                                                   & 0.29 / 0.16                                                              & 0.50 / 0.27                                                              & 1921.37 / 473.96                                                       & 0.14 / 0.37                                                               & 18.90 / 21.91                                                           & 0.58 / 0.71                                                             & \textbf{10.1 / 88.9}                                                                    \\
CFTE~\cite{CHEN2022DCC} / Ours           & 2.88 / 2.72                                                                   & 0.16 / 0.16                                                              & 0.26 / 0.27                                                              & 547.72 / 473.96                                                        & 0.33 / 0.37                                                               & 22.78 / 21.91                                                           & 0.73 / 0.71                                                             & \textbf{14.7 / 85.3}                                                                    \\
FNeVR~\cite{NEURIPS2022_8cc7e150} / Ours          & 3.14 / 2.72                                                                   & 0.31 / 0.16                                                              & 0.51 / 0.27                                                              & 1869.32 / 473.96                                                       & 0.15 / 0.37                                                               & 20.61 / 21.91                                                           & 0.64 / 0.71                                                             & \textbf{0.4 / 99.6}                                                                     \\
HiDeNeRF~\cite{li2023hidenerf} / Ours       & 4.09 / 2.72                                                                   & 0.28 / 0.16                                                              & 0.53 / 0.27                                                              & 1397.46 / 473.96                                                       & 0.26 / 0.37                                                               & 13.38 / 21.91                                                           & 0.38 / 0.71                                                             & \textbf{0.0 / 100.0}                                                                    \\
GPAvatar~\cite{chu2024gpavatar} / Ours      & 4.41 / 2.72                                                                   & 0.20 / 0.16                                                              & 0.32 / 0.27                                                              & 511.93 / 473.96                                                        & 0.45 / 0.37                                                               & 17.82 / 21.91                                                           & 0.58 / 0.71                                                             & \textbf{23.6 / 76.4}                                                                    \\ \hline
\end{tabular}
\vspace{-0.6em}
\end{table*}

\subsubsection{Subjective Quality}
Fig. \ref{fig7} shows the $50^{th}$, $150^{th}$ and $250^{th}$ frames of three particular talking face sequences with similar bit rates from different compression schemes. It can be noticed that at very low bit rates, the talking face videos reconstructed from the VVC codec exhibit severe blocking artifacts and facial distortions, whilst our proposed scheme can reconstruct talking face videos with vivid facial features and accurate facial motion. Regarding FOMM, FOMM, Face2FaceRHO and Face\_vid2vid algorithms, obvious compression artifacts can be easily perceived. In addition, at this low rate, the CFTE scheme cannot accurately represent expression and local motion (\textit{e.g.,} mouth motion and eye blinking) compared with our proposed scheme. As for the reconstructed video quality using FNeVR/HiDeNeRF/GPAvatar algorithms, it can be noticed that these NeRF-based methods can well achieve faithful identity reconstruction with precise head posture and facial expression. However, they suffer from poor background generation, i.e., their background cannot be accurately restored and is usually filled with black pixels, even though the reference texture have been provided. Different from these above compression schemes, our proposed method not only preserves global and local facial motion, but also reconstructs higher-quality face signal and its background at ultra-low bit rate.

In addition, a subjective test using the “two alternatives, forced choice” (2AFC) methodology was conducted to verify the subjective visual quality of our proposed scheme. More specifically, 15 participants are asked to choose the reconstructed video with better quality from a video pair at similar coding bit consumption. These participants are provided with two videos sequentially, where they can see either a video from method A or method B in right or left side. A total of 15 reconstructed sequences were selected in this 2AFC test, using the sequence generated by our proposed scheme and the corresponding reconstructed sequences from other algorithms to form a video pair. To avoid experimental bias, we shuffle all video pairs and display them randomly. As shown in Table \ref{table2}, when the pairwise reconstructed sequences have similar consumption of coding bits, these participants have a much higher percentage to choose our proposed method as the preferred video because of its better visual quality compared to other compression algorithms.

\begin{table*}[t]
\vspace{-1.5em}
\renewcommand\arraystretch{1.2}
\centering
\caption{Model complexity and coding efficiency comparisons among the proposed IFVC framework and different compression algorithms in terms of Params, MAdd and Time.} 
\label{table_efficiency_algorithm}
\centering
\resizebox{1\textwidth}{!}{ 
\begin{tabular}{cc|ccccccccc|c}
\hline
Codec                            & Measures                                                                   & \begin{tabular}[c]{@{}c@{}}VVC \\ (VTM10.0)\end{tabular} & \begin{tabular}[c]{@{}c@{}}FOMM \\ (NeurIPS'19)\end{tabular} & \begin{tabular}[c]{@{}c@{}}FOMM2.0 \\ (CVPR'21)\end{tabular} & \begin{tabular}[c]{@{}c@{}}Face\_vid2vid \\ (CVPR'21)\end{tabular} & \begin{tabular}[c]{@{}c@{}}Face2FaceRHO \\ (ECCV'22)\end{tabular} & \begin{tabular}[c]{@{}c@{}}CFTE\\  (DCC'22)\end{tabular} & \begin{tabular}[c]{@{}c@{}}FNeVR\\  (NeurIPS'22)\end{tabular} & \begin{tabular}[c]{@{}c@{}}HiDeNeRF\\  (CVPR'23)\end{tabular} & \begin{tabular}[c]{@{}c@{}}GPAvatar\\  (ICLR'24)\end{tabular} & \begin{tabular}[c]{@{}c@{}}IFVC \\ (Proposed)\end{tabular} \\ \hline
                                 & Params (M)                                                                 & \textbackslash{}                                         & 14.22                                                        & 14.15                                                        & 72.20                                                              & 25.85                                                             & 14.13                                                    & 14.22                                                         & 48.53                                                         & 51.69                                                         & 40.29                                                      \\
                                 & MAdd (G)                                                                   & \textbackslash{}                                         & 1.28                                                         & 1.06                                                         & 317.32                                                             & 4.09                                                              & 0.99                                                     & 1.28                                                          & 8.43                                                          & 60.51                                                         & 22.52                                                      \\
\multirow{-3}{*}{Encoder}        & Time (Sec)                                                                   & 1323.50                                                  & 21.07                                                        & 19.65                                                        & 18.83                                                              & 226.43                                                            & 16.40                                                    & 27.48                                                         & 43.43                                                         & 198.33                                                        & 44.50                                                      \\ \hline
                                 & Params (M)                                                                 & \textbackslash{}                                         & 45.58                                                        & 45.58                                                        & 53.10                                                              & 11.32                                                             & 43.89                                                    & 47.18                                                         & 87.46                                                         & 97.52                                                         & 44.59                                                      \\
                                 & MAdd (G)                                                                   & \textbackslash{}                                         & 53.64                                                        & 53.64                                                        & 178.28                                                             & 12.70                                                             & 54.82                                                    & 124.26                                                        & 245.02                                                        & 140.15                                                        & 117.49                                                     \\
\multirow{-3}{*}{Decoder}        & Time (Sec)                                                                   & 0.65                                                     & 14.27                                                        & 14.26                                                        & 20.09                                                              & 135.38                                                            & 13.23                                                    & 9.94                                                          & 109.38                                                        & 53.58                                                         & 177.49                                                     \\ \hline
                                 & \textbf{Params (M)}                                                        & \textbf{\textbackslash{}}                                & \textbf{59.79}                                               & \textbf{59.72}                                               & \textbf{125.30}                                                    & \textbf{37.16}                                                    & \textbf{58.02}                                           & \textbf{61.39}                                                & \textbf{135.99}                                               & \textbf{149.21}                                               & \textbf{84.88}                      \\
                                 & \textbf{MAdd (G)}                                                          & \textbf{\textbackslash{}}                                & \textbf{54.92}                                               & \textbf{54.70}                                               & \textbf{495.61}                                                    & \textbf{16.79}                                                    & \textbf{55.81}                                           & \textbf{125.54}                                               & \textbf{253.45}                                               & \textbf{200.67}                                               & \textbf{140.00}                    \\
\multirow{-3}{*}{\textbf{Total}}                                  & \textbf{Time (Sec)}                                                          & \textbf{1324.15}                                         & \textbf{35.34}                                               & \textbf{33.91}                                               & \textbf{38.92}                                                     & \textbf{361.81}                                                   & \textbf{29.63}                                           & \textbf{37.43}                                                & \textbf{152.81}                                               & \textbf{251.91}                                               & \textbf{221.98}                     \\\hline
\end{tabular}
}
\end{table*}

\begin{table}[t]
\vspace{-1.5em}
\renewcommand\arraystretch{1.2}
\centering
\caption{Total model complexity and coding efficiency comparisons regarding the proposed IFVC framework and its ablation versions in terms of Params, MAdd and Time.} 
\label{table_efficiency_abalation}
\centering
\resizebox{0.5\textwidth}{!}{ 
\begin{tabular}{cccc}
\hline
Measures                        & Params (M) & MAdd (G) & Time (Sec) \\ \hline
IFVC w/o Coarse-to-fine Motion  & 41.08      & 30.38    & 189.93     \\
IFVC w/o CSSFT-based Generation & 89.95      & 151.27   & 218.46     \\
IFVC w/o Eye-recalibration      & 84.88      & 140.00   & 184.87     \\
IFVC (Proposed)                            & 84.88      & 140.00   & 221.98     \\ \hline
\end{tabular}
}
\vspace{-1em}
\end{table}

\begin{table}[t]
\renewcommand\arraystretch{1.2}
\centering
\caption{Total model complexity and coding efficiency comparisons among the proposed IFVC framework and the pre-/post-manipulation algorithms for VVC in terms of Params, MAdd and Time.} 
\label{table_efficiency_frameworks}
\centering
\begin{tabular}{cccc}
\hline
\multirow{2}{*}{Measures} & \multicolumn{2}{c}{Face Manipulation Algorithm + VVC} & \multirow{2}{*}{\begin{tabular}[c]{@{}c@{}}IFVC\\  (Proposed)\end{tabular}} \\ \cline{2-3}
                          & Face\_vid2vid+VVC            & GPAvatar+VVC           &                                                                             \\ \hline
Params (M)                & 125.30                       & 149.21                 & 84.88                                                                       \\
MAdd (G)                  & 495.61                       & 200.67                 & 140.00                                                                      \\
Time (Sec)                & 1362.15                      & 1576.06                & 221.98                                                                      \\ \hline
\end{tabular}
\vspace{-1.5em}
\end{table}

\subsubsection{Interactive Face Video Coding}
Our proposed compression framework enjoys the best aspects of compact representation and semantic interpretation by characterizing highly-independent facial semantics from talking face frames. As such, it is of great benefits to realize controllable semantic interactivity or virtual character animation based on these facial semantics.
\begin{itemize}
\item{Controllable manipulation for friendly interactivity: By modifying the corresponding facial semantics at the decoder side, different interactive manners can be achieved in terms of eye blinking, mouth motion, head rotation, head translation and head location. As shown in Fig. \ref{fig8}, our proposed interactive face coding scheme enjoys great flexibility in controlling these semantics, where the shape and motion of the eyes and the mouth, as well as the head posture can all be separately controlled. Such a compression mechanism can well support future applications that require both high coding performance and friendly interactivity. }%
\item{Virtual character animation/stylization for privacy protection: To enable the protection of user privacy in talking face communication, the proposed IFVC system can treat a virtual character as the key-reference frame and animate it with compact facial semantics at the decoder side. More specifically, virtual characters can be selected by end users, and the decoded facial semantics from the coded bitstream can be used to animate the virtual face character with accurate pose and expression. Fig. \ref{fig9} shows the robustness of our proposed algorithm using virtual characters with very different features. As shown in these samples, the proposed IFVC framework can animate any cross-identity face character with compact facial semantics to produce face images with accurate eye/mouth motion and head posture, providing great possibility for virtual live entertainment or face video stylization. 
It should be mentioned that virtual character animation results in Fig. \ref{fig9} may exist unpleasant visual effects like jittering due to the domain gap between virtual character and real-face signal. In particular, the adopted WM3DR~\cite{Zhang2021WeaklySupervisedM3} that is only trained at real-face datasets cannot well ensure the accurate parameter regression and mesh reconstruction of the virtual character, which will influence the subsequent mesh-based motion estimation and face video animation.
} %
\end{itemize}

\subsubsection{Ablation Study} 
To further demonstrate the effectiveness of various components and optimization schemes in our proposed algorithm, we have carried out the ablation studies in architecture modules, mouth parameter selection and optimization functions. As shown in Fig. \ref{ablation} (a), we first execute the ablation experiment in IFVC architecture designs regarding coarse-to-fine motion estimation, CSSFT-based frame generation and eye recalibration. In particular, for the ablation of coarse-to-fine motion, we remove this strategy and directly use the mesh-approximated flow as motion guidance. In addition, regarding the ablation of frame generation, we remove the CSSFT design and directly employ the U-NET architecture to reconstruct frames. In terms of eye recalibartion, the OpenFace library is abandoned to label the motion change in eye parts, and the extracted parameter dimension is changed from 14 to 13. Experimental results demonstrate every component and mechanism of the proposed framework can improve the RD performance, especially for the coarse-to-fine motion estimation module. 

In addition, Fig. \ref{ablation} (b) provides the ablation experiment in different vector dimension selections for mouth motion coefficients. Experimental results demonstrate when the dimension of mouth motion coefficients is set at 6, the overall RD performance for different quality measures in terms of DISTS, FVD, MANIQA and PSNR could be optimal. As shown in Fig. \ref{ablation} (c), when all loss functions are employed to jointly optimize our proposed IFVC model, the RD performance can be all improved in terms of DISTS, FVD, MANIQA and PSNR.

\subsubsection{Model Complexity}
To demonstrate the trade-offs between performance gains and increased model complexity, we have further provided different comparisons. In particular, we have first compared our proposed IFVC algorithm with other generative compression algorithms. In addition, we have compared the proposed IFVC with its ablation versions. Finally, we have compared our proposed framework with the existing interactive framework. Herein, we mainly focus on the number of model parameters (Params/M), number of operations measured by multiply-adds (MAdd/G) and actual inference time (Time/Sec). In particular, the inference process of coding efficiency is executed on Tesla-A100 with 15 core CPUs (Intel(R) Xeon(R) Platinum 8369B CPU @ 2.90GHz), and the testing sequence has 250 frames at the resolution 256$\times$256. Besides, the actual inference time at the encoder and decoder sides are both averaged with 5 different QPs (QPs=32, 37, 42, 47 and 52). It should be worth mentioning that the VVC codec relies on CPUs to compress videos, while GFVC algorithms can use GPU for acceleration. 

As illustrated in Table \ref{table_efficiency_algorithm}, the proposed IFVC algorithm is able to achieve good trade-offs among promising RD results, interactive visual applications and reasonable model complexity compared with the existing GFVC algorithms. Specifically, although our proposed IFVC is more complex and time-consuming than FOMM/FOMM2.0/CFTE, it is able to provide better RD performance and more diverse functionalities in interaction. Besides, in comparison with these Face\_vid2vid/Face2FaceRHO/FNeVR/HiDeRF/GPAvatar algorithms that can also achieve interactive functions, our proposed IFVC has lower model complexity. In addition, Table \ref{table_efficiency_abalation} shows that each component/mechanism in the proposed framework can improve RD performance without introducing too much model complexity. In particular, although the coarse-to-fine motion estimation mechanism can greatly reduce the model complexity, its RD performance drops significantly without any gains. As shown in Table \ref{table_efficiency_frameworks}, the overall coding efficiency and model complexity of our proposed IFVC framework are relatively superior compared with the traditional methods manipulating the face in the transmitted video via pre-processed or post-processed algorithms.

\section{Conclusions}
In this work, we propose a comprehensive IFVC solution to meet the exponential increase in the demand for low-delay, high-efficiency and interactivity enhanced face video compression, based on which a series of interactive applications for communication (\textit{e.g.,} metaverse based on virtual character, virtual art and live entertainment) are made possible. The proposed IFVC system is based on a core IDI scheme, which can realize the internal conversion of the input 2D face frame to 3D facial mesh representations. The IDI scheme greatly increases the degree-of-freedom for interactivity, and produces semantically meaningful bitstream that is featured with compact, controllable and flexible representations. Experimental results show that our proposed IFVC scheme is capable of achieving the superior RD performance and subjective quality for talking face video compression compared with the state-of-the-art VVC standard and the latest generative compression schemes. More importantly, validations also show that the interactivity can be supported with accurate manipulation of facial expression and head posture, whilst interactivity can be seamlessly integrated with the decoding process.

\bibliographystyle{IEEEtran}
\bibliography{bare_jrnl}

\begin{IEEEbiography}
[{\includegraphics[width=1in,height=1.25in,clip,keepaspectratio]{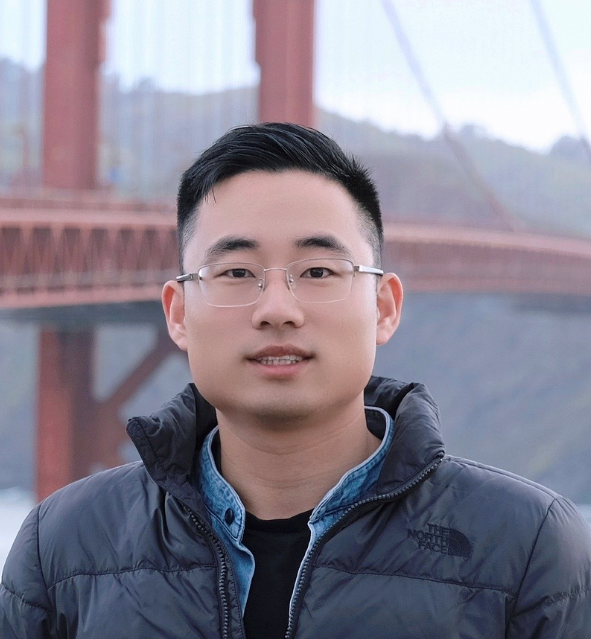}}]{Bolin Chen} (Member, IEEE) received the B.S. degree in communication engineering from Fuzhou University in July of 2020 and the Ph.D. degree in computer science from City University of Hong Kong in February of 2025, respectively. He has submitted more than 30 technical proposals to ISO/MPEG and ITU-T standards, and published more than 20 refereed journal articles/conference papers. His research interests include video compression, quality assessment and multimedia processing.
\end{IEEEbiography}

\begin{IEEEbiography}
[{\includegraphics[width=1in,height=1.25in,clip,keepaspectratio]{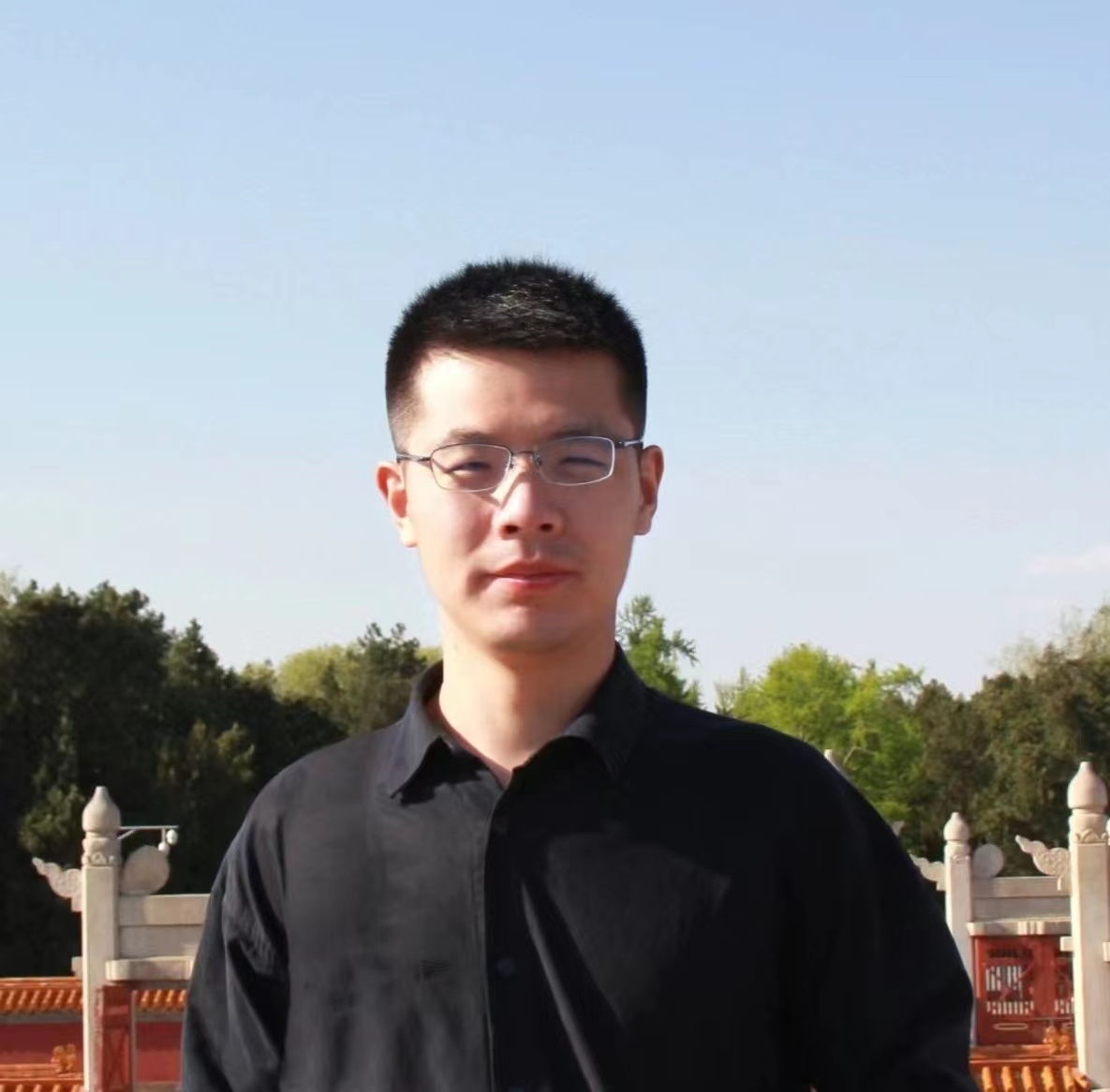}}]{Zhao Wang} received the B.S. degree in software engineering from Fudan University in 2014 and the Ph.D. degree in computer science from Peking University in 2019. From 2019 to 2022, he was an Algorithm Expert with the DAMO Academy, Alibaba Group. He is currently an Associate Researcher with the School of Computer Science, Peking University. He has proposed more than
40 technical proposals to video coding standards, including H.266/VVC, JVET-NNVC, MPEG-VCM, and AVS. He has published more than 30 articles. His research interests include video compression, generation and analysis. He received the Best Student Paper Award from the IEEE ICIP 2018.
\end{IEEEbiography}

\begin{IEEEbiography}[{\includegraphics[width=1in,height=1.25in,clip,keepaspectratio]{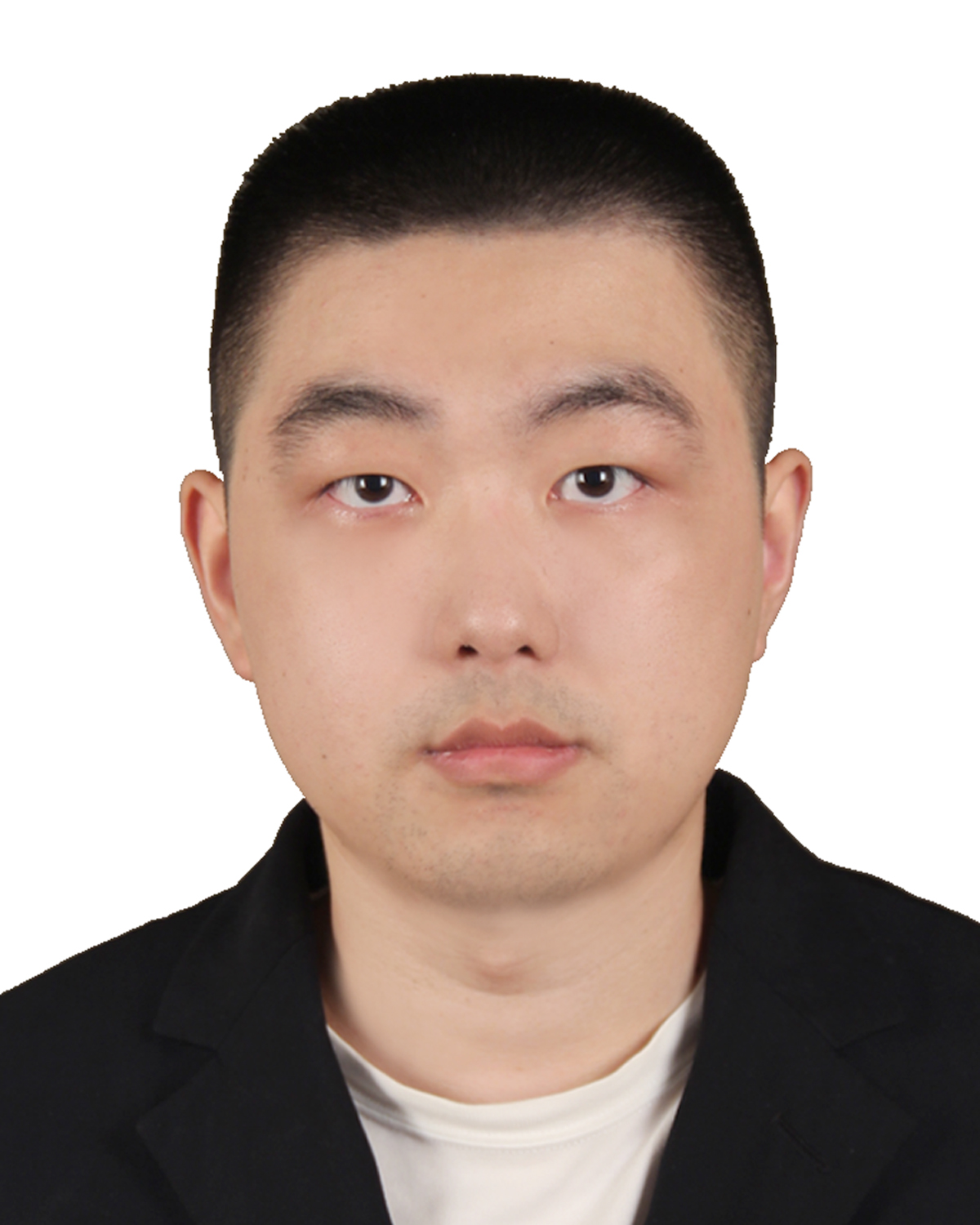}}]{Binzhe Li} received the B.S. degree in automation from Northeastern University in 2017, the M.S. degree in Pattern Recognition and Intelligent System from Huazhong University of Science and Technology in 2020 and the Ph.D. degree in computer science from City University of Hong Kong in 2024. His current research interests include image/video compression. 
\end{IEEEbiography}

\vspace{-3em}
\begin{IEEEbiography}
[{\includegraphics[width=1in,height=1.25in,clip,keepaspectratio]{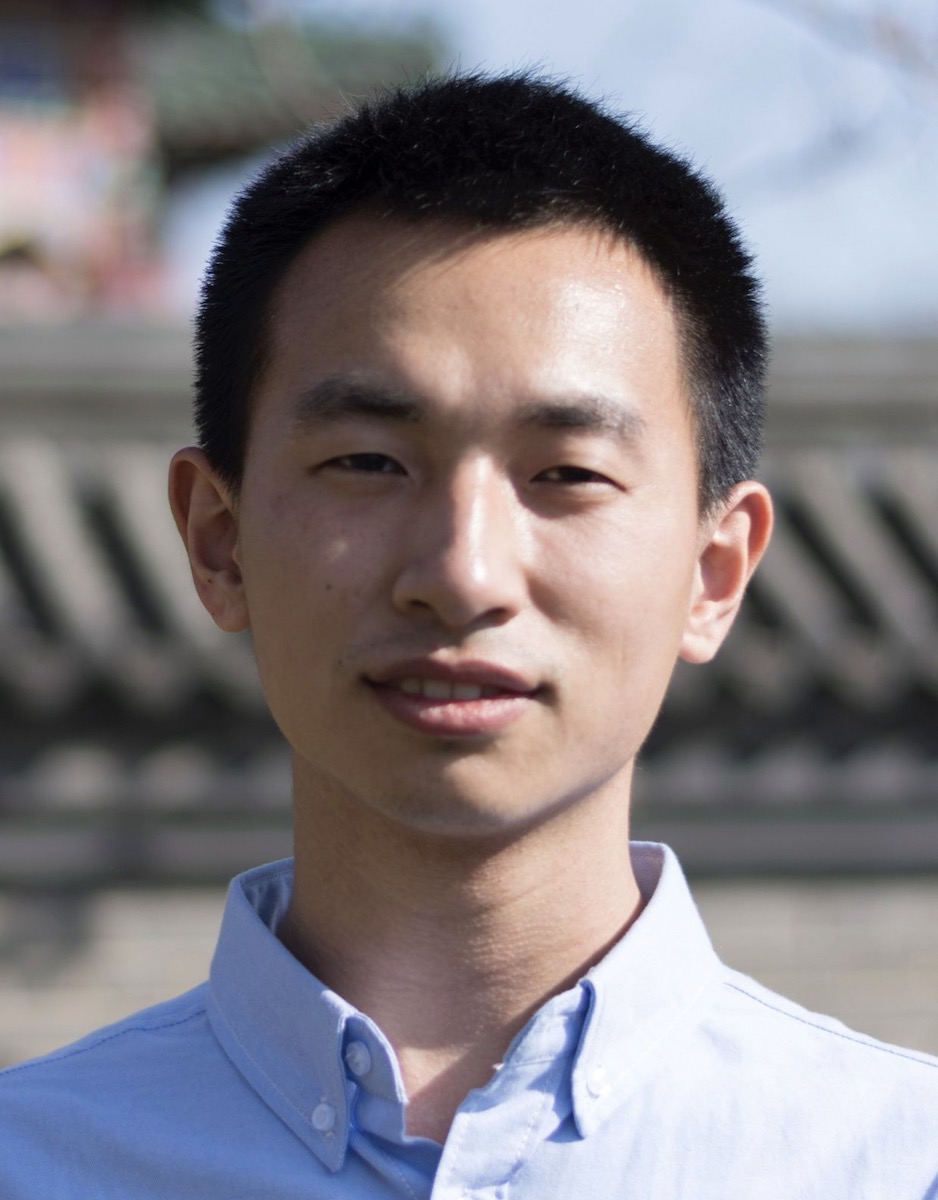}}]{Shurun Wang} received the bachelor’s degree and master’s degree in applied mathematics and computer science and technology of Peking University in 2016 and 2019, respectively. He pursues the Ph.D. degree of computer science with the Department of Computer Science, City University of Hong Kong in 2023. He is currently a senior algorithm engineer in Alibaba Group. His main research interests include the processing and compression of image/video towards machine vision, deep learning feature compression and Artificial Intelligence Generated Content (AIGC).
\end{IEEEbiography}

\vspace{-3em}
\begin{IEEEbiography}[{\includegraphics[width=1in,height=1.25in,clip,keepaspectratio]{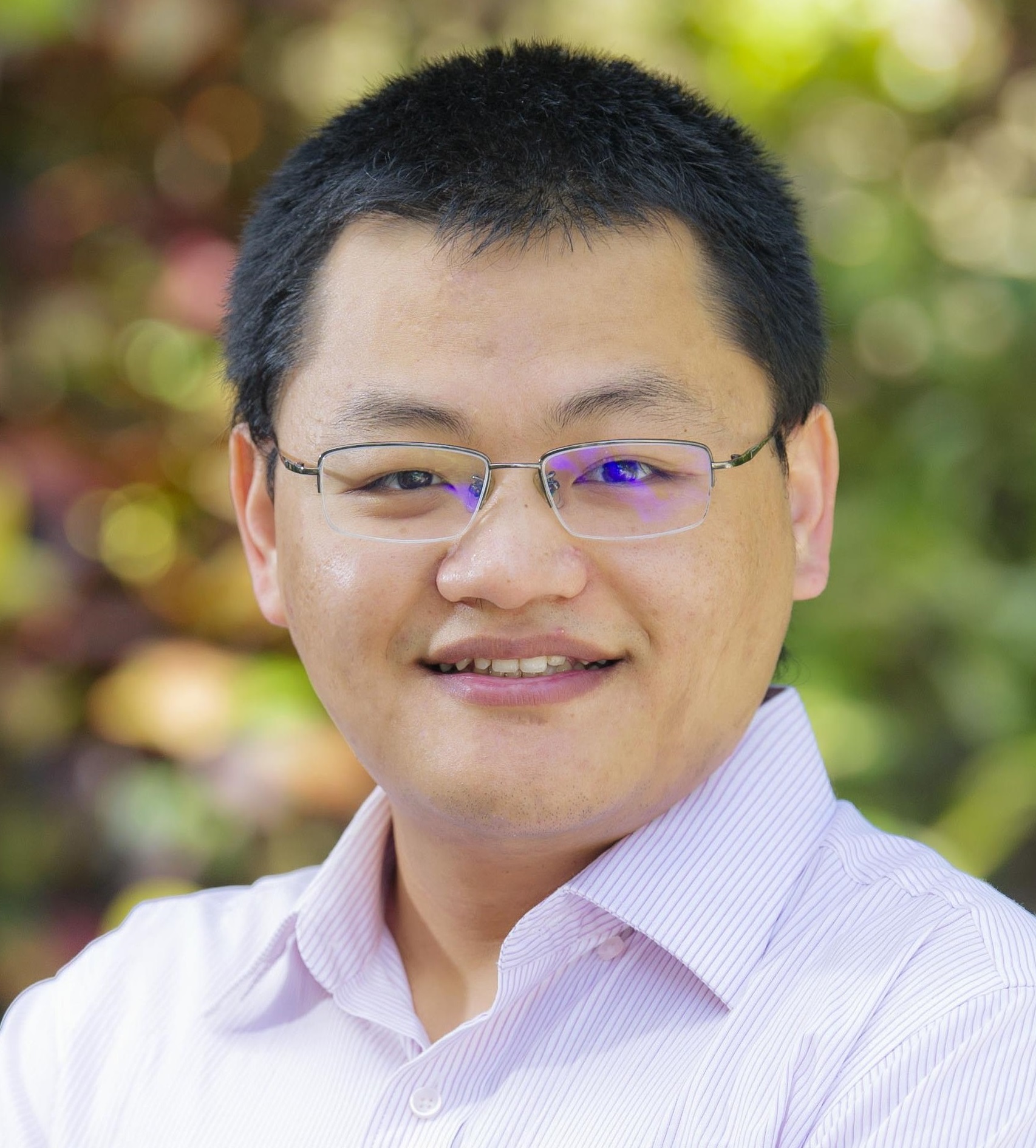}}]{Shiqi Wang} (Senior Member, IEEE) received the B.S. degree in computer science from the Harbin Institute of Technology in 2008 and the Ph.D. degree in computer application technology from Peking University in 2014. From 2014 to 2016, he was a Post-Doctoral Fellow with the Department of Electrical and Computer Engineering, University of Waterloo, Waterloo, ON, Canada. From 2016 to 2017, he was a Research Fellow with the Rapid-Rich Object Search Laboratory, Nanyang Technological University, Singapore. He is currently an Associate Professor with the Department of Computer Science, City University of Hong Kong. He has proposed more than 50 technical proposals to ISO/MPEG, ITU-T, and AVS standards, and authored or coauthored more than 300 refereed journal articles/conference papers. His research interests include video compression, image/video quality assessment, and image/video search and analysis. He received the Best Paper Award from IEEE VCIP 2019, ICME 2019, IEEE Multimedia 2018, and PCM 2017. His coauthored article received the Best Student Paper Award in the IEEE ICIP 2018. He was a recipient of the 2021 IEEE Multimedia Rising Star Award in ICME 2021. He serves as an Associate Editor for \textsc{IEEE Transactions on Circuits and Systems for Video Technology}, \textsc{IEEE Transactions on Image Processing}, \textsc{IEEE Transactions on Multimedia} and \textsc{IEEE Transactions on Cybernetics}. 
\end{IEEEbiography}

\vspace{-3em}
\begin{IEEEbiography}[{\includegraphics[width=1in,height=1.25in,clip,keepaspectratio]{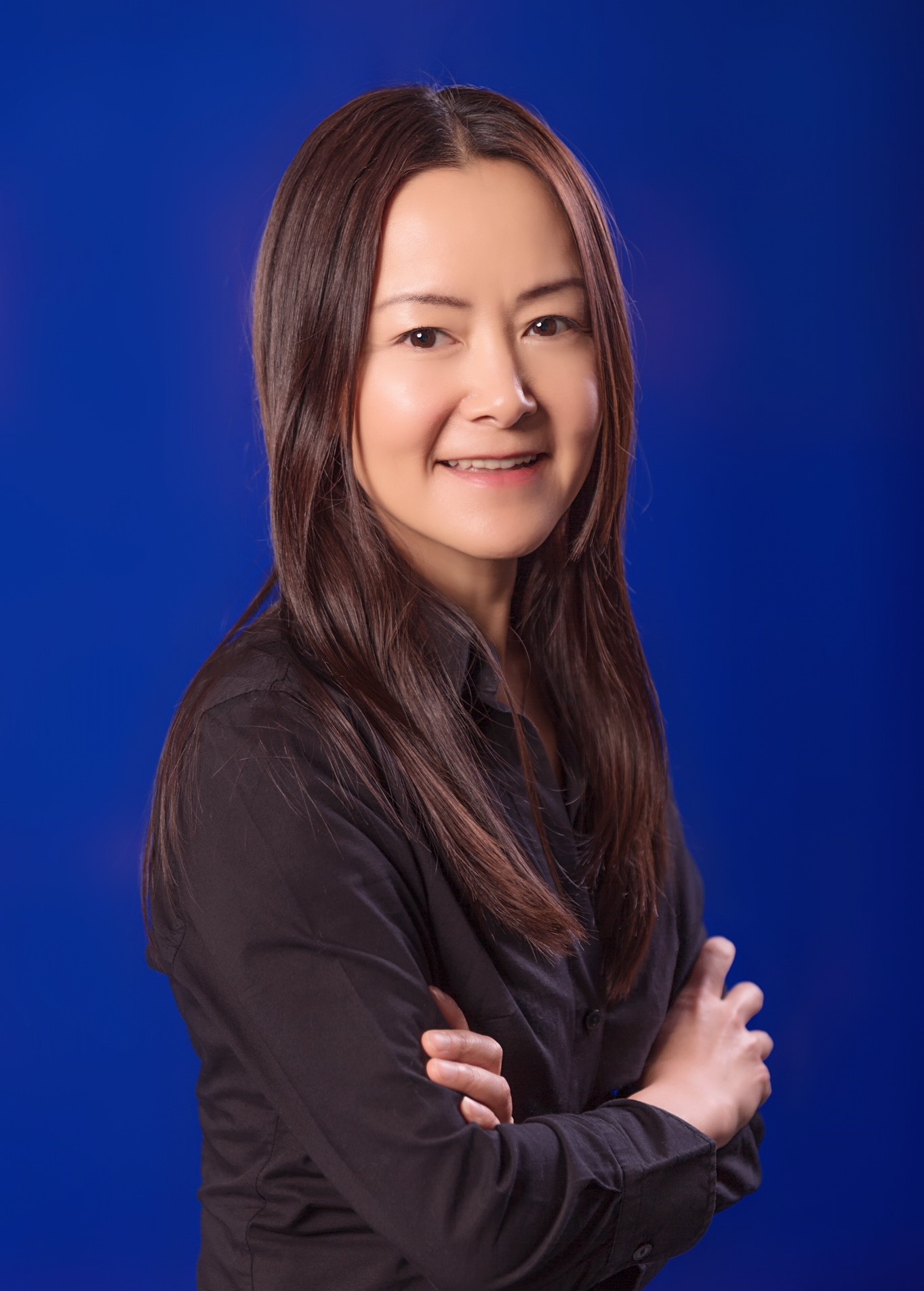}}]{Yan Ye} (Senior Member, IEEE) received the B.S. and M.S. degrees in electrical engineering from the University of Science and Technology of China in 1994 and 1997, respectively, and the Ph.D. degree in electrical engineering from the University of California at San Diego, in 2002. She is currently the head of Video Technology Lab at Alibaba Damo Academy, Alibaba Group U.S., Sunnyvale, CA, USA, where she oversees multimedia standards development, video codec implementation, and AI-based video research. Prior to Alibaba, she was with the Research and Development Labs, InterDigital Communications, Image Technology Research, Dolby Laboratories, and Multimedia Research and Development and Standards, Qualcomm Technologies, Inc. She has been involved in the development of various video coding and streaming standards, including H.266/VVC, H.265/HEVC, scalable extension of H.264/MPEG-4 AVC, MPEG DASH, and MPEG OMAF. She has published more than 60 papers in peer-reviewed journals and conferences. Her research interests include advanced video coding, processing and streaming algorithms, real-time and immersive video communications, AR/VR, and deep learning-based video coding, processing, and quality assessment. 
\end{IEEEbiography}

\end{document}